\documentclass[manuscript,screen,camera-ready,acmlarge]{acmart}
\usepackage{booktabs}
\usepackage{makecell}
\usepackage{xcolor}
\usepackage{tabularx}
\usepackage{tabularray}
\usepackage{longtable}

\AtBeginDocument{%
  \providecommand\BibTeX{{%
    \normalfont B\kern-0.5em{\scshape i\kern-0.25em b}\kern-0.8em\TeX}}}

\copyrightyear{2022}
\acmYear{2022}
\acmDOI{10.1145/3533708}

\begin{document}

\title{Federated Learning for Healthcare Domain - Pipeline, Applications and Challenges}


\author{Madhura Joshi}
\authornote{Both authors equally contributed to this research.}
\email{m.joshi@saama.com}
\author{Ankit Pal}
\authornotemark[1]
\email{ankit.pal@saama.com}
\affiliation{%
  \institution{Saama AI Research}
  \city{Chennai} 
  \country{India}
}

\author{Malaikannan Sankarasubbu}
\affiliation{%
  \institution{Saama AI Research}
  \city{Chennai} 
  \country{India}
}
\email{malaikannan.sankarasubbu@saama.com}

\begin{abstract}
Federated learning is the process of developing machine learning models over datasets distributed across data centers such as hospitals, clinical research labs, and mobile devices while preventing data leakage. This survey examines previous research and studies on federated learning in the healthcare sector across a range of use cases and applications. Our survey shows what challenges, methods, and applications a practitioner should be aware of in the topic of federated learning. This paper aims to lay out existing research and list the possibilities of federated learning for healthcare industries.
\end{abstract}

\ccsdesc[500]{Security and privacy}
\ccsdesc[500]{Computing methodologies~Artificial intelligence}

\keywords{federated learning, GDPR, transfer learning}

\maketitle

\section{\textbf{Introduction}}
In the last few years, digital healthcare data has grown significantly. At the same time, recent breakthroughs in deep learning (DL) have been used in a variety of current medical data processes, including automatic disease diagnosis \cite{Liu2020,Rajan2019}, classification, biomedical data analysis, Question Answering in the medical domain \cite{pmlr-v174-pal22a} and segmentation \cite{Li2015,Menze2015}. These methods hold immense promise and innovation in this field. In the coming future, the advancement of these methods will refine health care systems and improve medical practices worldwide.

Diagnostic tools, machine learning (ML) based healthcare solutions, and models must be exposed to a wide variety of cases and data that cover a full range of possible anatomies to capture more informative patterns in the medical data. It is well known that data from a single source can be significantly biased by the equipment, demographics, and acquisition protocol. Therefore, training a model on data from a single source would skew its prediction performance towards the population. Moreover, it is computationally expensive and time-consuming.

Training models in a parallelized manner \cite{Dean2012} and within small batches can mitigate a few of these challenges. However, though this approach addresses computation challenges, it does not necessarily preserve the privacy of data. 
Clinical research often involves studies from a large amount of data collected from various sources. Health institutions, individuals, insurance companies, and the pharmaceutical industry all have access to medical data. Furthermore, each institution may be linked to a unique collection of stakeholders. These data are often sensitive and cannot be aggregated or accessed.

Access to a large amount of high-quality medical data is possibly the most crucial factor for enhancing Machine Learning (ML) applications in the healthcare domain. However, security and privacy issues of healthcare data have raised broad ethical and legal concerns in recent years, given the sensitive nature of health information \cite{Adam2007}.

The assembly and transmission of these datasets are ethically and legally required to protect patient privacy. Most healthcare centers, laws at the country level, and regulatory bodies, e.g., the General Data Protection Regulation (GDPR) and the Health Insurance Portability and Accountability Act (HIPAA), have passed new laws that control sharing data while preserving user security and privacy\cite{Yang2019,Li2020}. Moreover, information and control about medical data storage, transmission, and usage are central to patient rights.

The sensitive and distributed nature of EHR (Electronic Health Records) in real-world scenarios simulates a need for an effective mechanism to learn from data residing in health-related institutions, hospitals while accounting for data privacy. This motivates us to examine the potential and value of federated learning for the healthcare domain. Federated Learning is an advanced distributed learning technique that leverages datasets from various universities without explicitly centralizing or sharing the training data. \cite{Li2020FederatedLC,10.1145/3298981}.

Federated learning provides many advantages as compared to centralized learning. It enables training a global model from distributed data. This method also focuses on preserving data privacy by only sharing mathematical parameters and metadata while keeping the actual data as secure as possible and preventing attacks and tracebacks.

The global ML model is distributed to each client site, where an instance is trained locally. The updates from locally trained instances are then aggregated at regular intervals to improve the global model. The updated global model is then sent back to the local devices, where the learning continues. These steps are repeated until a particular convergence threshold is satisfied or lasts for a long time to improve the deep learning model continuously.

The parameters and metadata sharing depend on many factors such as use cases, data management and regulation, business agreement and protection, pipeline, and infrastructure. In this article, we provide an overview of the federated learning approach in the healthcare domain. Our contribution is as follows
\begin{itemize}
    \item We demonstrate the components of the federated learning setup and discuss the communication architecture and building blocks of a federated learning system.
    \item We examine the various challenges that a federated learning setup faces in terms of privacy, data, and communication in the healthcare system.
    \item We survey existing works on federated learning in the health sector and propose a comprehensive list of applications classified into prognosis, diagnosis,  and clinical workflow.
\end{itemize}

The structure of the paper is as follows; the first section describes the components of a federated learning setup as well as a federated learning pipeline. Section 2 discusses the challenges and concerns of federated learning in the healthcare area. Section 3 concludes with a study of federated learning applications in the health care area and the tools required for implementation.

\section{\textbf{PIPELINE}}
\subsection{\textbf{Architecture of Communication}}
\subsubsection{Centralized}

Centralized federated learning architecture is the most commonly used architecture.
In this setting, data flows in asymmetric nature. A central server is responsible for aggregating the information (e.g., local models) from the data owners, coordinating with all the participating client devices, and sending back the model updates. Client devices usually communicate only with the server; hence the server acts as a system bottleneck. However, Single-point failure is one of the drawbacks in a centralized setting. 

The following is the architecture of centralized federated learning: first, a global model is sent to client devices. Second, the global model is being updated with client data. Third, the revised model is sent back to the server. Finally, local models are averaged on the server to create a new global model. Furthermore, this process is repeated. The architecture of centralized federated learning is shown in figure 1.

Google uses the centralized architecture design in their Android keyboard setting. The central server collects the local model information from users' mobile devices and updates the global model, and later the updates are sent back to the users for inferences. FedAvg is one such example of a centralized framework\cite{McMahan2017}.

\begin{figure}[h]
  \centering
  \includegraphics[width=10cm]{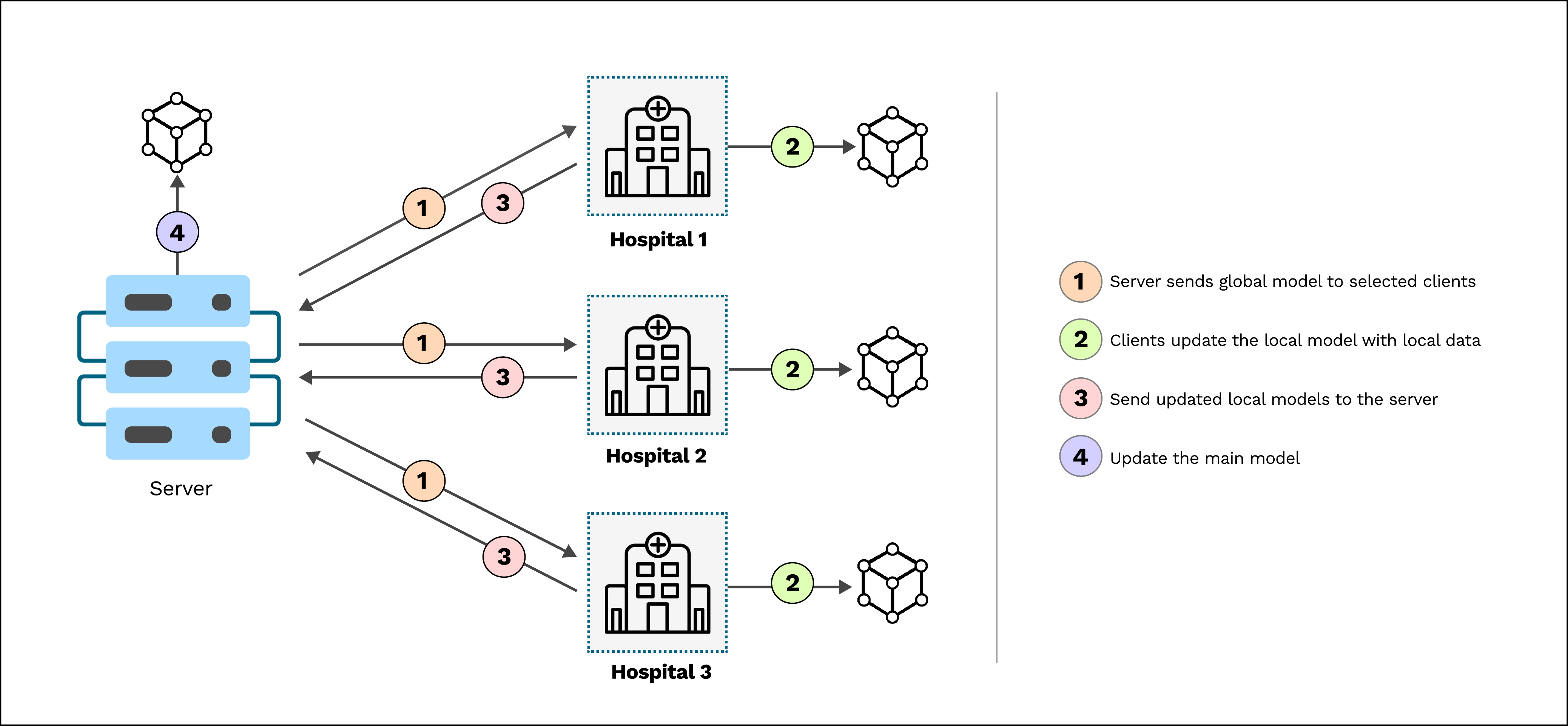}
  \caption{Illustration of centralized architecture in federated learning}
  \Description{A woman and a girl in white dresses sit in an open car.}
\end{figure}

\subsubsection{Decentralized}
In the decentralized, federated learning architecture, the client devices can communicate to train a global model and update it directly without any central server. A decentralized setup prevents the single point failure issue. SimFL is a federated, decentralized framework \cite{Li2020FederatedLC}.

The following is the architecture of decentralized, federated learning: first, the local gradients are updated. Second, these gradients are sent to selected parties. Third, the model is updated with local data and gradients. Lastly, the updated model is sent to other parties, and the process continues. The architecture of decentralized, federated learning is shown in figure 2. This method is usually preferred because it involves exchanging of local models which are aggregations of large quantity of data. However, decentralized federated learning works on the concept of mutual trust between the users which leads to few drawbacks.The question arises when one has to use DFL in a single sided trust environment. A decentralized learning algorithm called Online Push Sum (OPS)\cite{he2020central} addresses this challenge.It involves a rigorous regret analysis, tested and compared with other algorithms like Decentralized Online Gradient (DOG) and Centralized Online Gradient (COG).

Blockchain is an excellent example of a decentralized platform \cite{Zheng2018}. Another example of a similar design is the decentralized cancer diagnostic system amongst hospitals/medical institutions. Each hospital distributes the local model that has been trained with patient data and acquires the global model for future diagnosis \cite{Sarma2021}.

\begin{figure}[h]
  \centering
  \includegraphics[width=8cm]{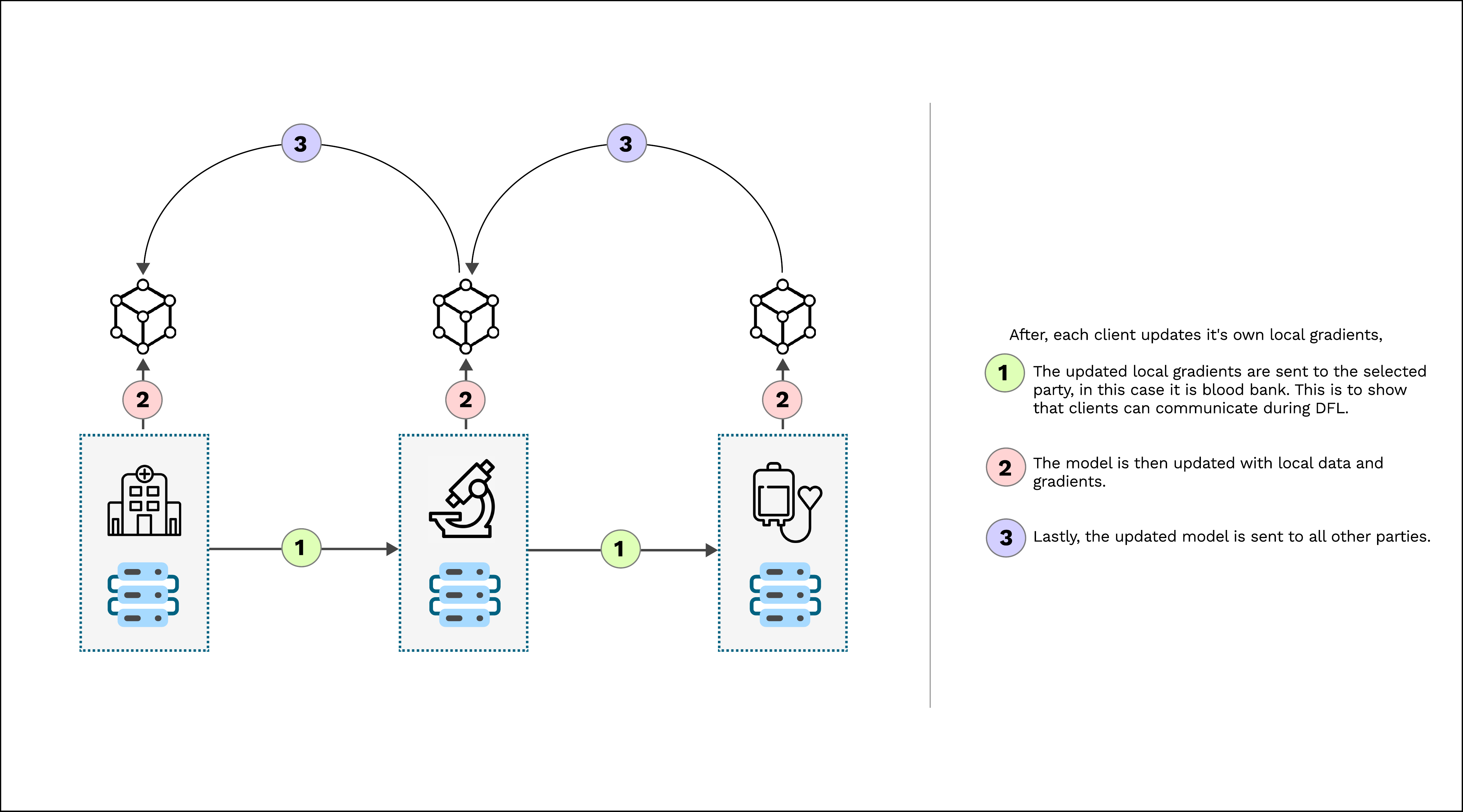}
  \caption{Illustration of decentralized architecture in federated learning}
  \Description{A woman and a girl in white dresses sit in an open car.}
\end{figure}

\subsection{\textbf{Scale of Federated Learning system}}
\subsubsection{Cross-Silo}
In a Cross-Silo federated learning setting, small numbers of organizations or data centers (e.g., medical or financial) train a global model with a large amount of data along with computation power. In an environment where organizations are prohibited from sharing their data due to data privacy and security reasons, cross-silo federated learning comes into the picture. The computation and storage capacity of the cross-silo setting is high on a relatively small scale, while the stability is low compared to cross-device.

\subsubsection{Cross-Device}
In a Cross-Device federated learning setting, there are a scalable number of clients with small amounts of data compared to a cross-silo setting where there are only a few clients with large amounts of data. As a result, the cross-device system develops models for large-scale dispersed data inside the same application \cite{lo2021architectural}; the clients can be enabled and disabled as per the requirement and are usually mobile and IoT devices. The system should be powerful enough to manage many devices and handle common failures such as unstable connections, power failure, etc.

\subsection{\textbf{Federated Server}}

The central server serves as a manager in a cross-device environment. It oversees the communication between the client devices, the server, and the global machine learning model. At the beginning of the federated learning training process, the client starts a server training service; a login request is required from each client-side to join the server training session.

The server will check the credentials to validate the request and allow the clients to join the session only in a successful authentication. The server session decides the minimum and a maximum number of client devices that need to be added to start the training process.

In a decentralized environment, all devices interact directly with one another and participate equally in global model training. The server, in this case, is all of the local devices. Practical Federated Gradient Boosting Decision Trees \cite{Li2019} is an example of such a framework, in which each device trains decision trees sequentially, and the final model is the sum of all the trees. Designing a completely decentralized FLS with appropriate communication overhead is difficult.

Each local model training happens on the client's side, so the server does not need access to the model training data. Moreover, local models only share the model updates instead of the actual copy of data. The client's devices decide the number of epochs to run during each round of training. Typically, the computing is used for model training and aggregation, while the communication is used to exchange model parameters.

Once the server receives all the model updates from all the participating devices, it performs the model aggregation based on the selected aggregation algorithm and gets the overall updated global model. This process completes the single round of the Federated Learning session. Furthermore, the next round starts and training continues until the maximum number of rounds on the server is completed. The aggregation technique aids in the gathering of model updates from local models of the devices. Listed below are some popular application-oriented aggregation algorithms in detail. 

\begin{itemize}
    \item \textbf{FedAvg Algorithm}
    FedAvg is the first and most widely used Federated learning algorithm proposed by Google \cite{McMahan2017}. It distributes training data across mobile devices and trains a shared model by combining locally calculated updates. The FedAvg algorithm is a robust approach experimented on five different architectures, four kinds of datasets, and combined stochastic gradient descent on each client with a server that performs model averaging. The algorithm can be written as 
    
    \begin{equation}
f(w)=\sum_{k=1}^{K} \frac{n_{k}}{n} F_{k}(w) \quad \text { where } \quad F_{k}(w)=\frac{1}{n_{k}} \sum_{i \in \mathcal{P}_{k}} f_{i}(w)
\end{equation}

Where $\mathcal{P}_{k}$ the set of indexes of data points on client k, distributed uniformly at random, with ${n_{k}}$ = $|Pk|$ and $w$ are model parameters. The focus of the proposed research is on the optimization's non-IID data distributions and imbalanced properties, as well as the nature of the communication constraints.

\item \textbf{FedMa Algorithm}
FedMA \cite{Wang2020} is a federated learning method for current neural network designs such as Convolutional Neural Networks (CNNs) and LSTMs. FedMA builds the shared global model layer by layer by matching and averaging hidden components with comparable feature extraction signatures (i.e. channels for convolution layers; hidden states for LSTM; neurons for fully connected layers).

\item \textbf{FedNAS Algorithm}
One of the major issues on working with Non-I.I.D data is that it requires the developers to design and choose multiple architectures that tunes the hyper parameters and fits the scattered data. This is design process is expensive due to large number of communication rounds and computational burden. FedNAS - Federated Neural Architecture Search \cite{DBLP:journals/corr/abs-2004-08546} is an algorithm that improves the model accuracy by automating the manual design process. The algorithm helps scattered workers to collaboratively search for a better architecture. 

\item \textbf{FedGKT Algorithm}
The fact that federated learning ensures privacy and confidentiality is one of its most appealing features. There is, however, a limit to the computation capabilities of the edge nodes. Convolutional neural networks are used to boost the model's performance and accuracy, however when employed with FL restrictions, large models might burden edge nodes and increase communication costs. A group knowledge transfer training technique intends to reduce the need for edge compute while keeping edge training inexpensive by lowering communication bandwidths for training with CNNs at the edge. FedGKT \cite{DBLP:journals/corr/abs-2007-14513} requires 9 to 17 times less computing resources on edge devices than FedAvg. In comparison to FedAvg, the authors claim that this algorithm can achieve the same or slightly higher accuracy. 

\item \textbf{Secured Weighted Aggregation}
The proposed algorithm \cite{Guo2021} is relying on homomorphic encryption - cryptosystem for calculating the client’s weights in a privacy-preserving manner. The algorithm adopts a Zero-Knowledge Proof(ZKP) based verification scheme to prevent the central servers and clients from receiving fraudulent messages from each other. This is the first aggregation algorithm that deals with Data disparity and fraudulent messages

\item \textbf{Secure Federated matrix factorization}
The authors of this research offer a Federated Matrix Factorization Framework to address the problem of disclosing data through gradients FedMF \cite{Chai2019}. Distributed machine learning and homomorphic encryption techniques are used in the system. FedMF is a user-level distributed matrix factorization framework that allows the model to be learned by uploading the gradient information to the server rather than the raw preference information. The authors utilize homomorphic encryption to improve distributed matrix factorization and boost security since gradient information can leak user information to some extent.

\item \textbf{ Inprivate Digging}
The authors concentrate on a tree-based data mining concept and provide privacy-preserving approaches for two of the most popular tasks: Regression and Binary Classification, in which individual data owners can train locally in a differentially private way \cite{Zhao2018}. They created and implemented a privacy-preserving Gradient Boosting Decision Tree (GBDT) system that allows several regression trees trained by distinct data owners to be safely aggregated into an ensemble without the need of a third party.

\item \textbf{ Federated Forest}
The authors provided a system that allows random forests to be trained in a vertical FL scenario \cite{Liu2020}. Throughout the development of each node, the party with the related split feature is in charge of splitting the samples and exchanging the results. To safeguard privacy, they encrypt the data that is sent. Their method is equally accurate as of the non-federated one. Both classification and regression tasks are supported by the model.

\end{itemize}

\subsubsection{Model Evaluation}

In traditional centralized machine learning, evaluation metrics are used to assess the performance and measure the quality of the model. Evaluation metrics mainly include accuracy, precision, and recall, etc. Accuracy is a fraction of the correct samples to all samples. Precision is the fraction of actual positive samples among the positive samples, while recall is the fraction of actual positive samples among the samples from true positive or false negative.

In a federated learning setup, the above metrics are insufficient to evaluate the performance of the models.  In this setting, it is intended to identify the evaluation metrics for both qualitative and quantitative methods. Training speed, performance, and the quantity of data transferred are utilized as assessment measures in existing federated learning studies. An FL model is evaluated by examining the aggregated model after being assigned to each clients' local evaluation dataset. Following then, the server shares each client's performance with it, aggregating the local results to produce global assessment metrics.

FedEval \cite{chai2020fedeval} is an evaluation framework for FL systems. It introduces the “ACTPR” model, i.e., using accuracy, communication, time consumption, privacy, and robustness as its evaluation targets.

\subsection{\textbf{Federated Client}}

In a federated learning setting, the client devices are the hardware components that perform the local model training and loss minimization The main objective is to train local device models on their private data and share the updates with the federated server where the aggregation is performed. In a decentralized federated learning system, the client devices communicate among themselves without the interference of the central server. Typical clients in federated learning settings could be smartphones, IoT devices, or organizations such as hospitals. Each client has its specific private training dataset and its local model. 

A study in \cite{lo2021architectural} describes the client management patterns that control the local devices' information and their connection with the central server. The client registry is the initial component, and it controls the information about the participating devices. The second component, the client selector, selects the devices for the federated training task. To improve the training model's performance and efficiency, the third component, client cluster, clusters client devices depending on particular parameters (e.g., data distribution, available resources).
\noindent
\subsubsection{Data Partitioning}

Federated learning systems are commonly classified as horizontal, vertical, or hybrid based on how data is spread through the sample and feature spaces. 

\noindent

\begin{itemize}
    \item \textbf{Horizontal Federated Learning}

\begin{itemize}
    \item \textbf{Concept} 
Horizontal federated learning involves data that is shared horizontally which involves datasets sharing the same feature space but being different in samples \cite{10.1145/3298981}. Each group in horizontal FL has access to the entire feature set and labels, allowing them to train their local model using their dataset. After that, all of the parties exchange their model updates with an aggregator and then generate a global model by combining, for example, the model weights obtained from different parties. \cite{xu2021fedv} An example of horizontal data split in federated learning setup is shown in figure 3. Assuming there is a blood bank laboratory having details of the patients stored as groups in a database collected under Name, Age, and Blood Person features. In figure 3, persons 1, 2, and 3 represent groups of patients’ data samples sharing the same features – Name, Age and Blood Group. Suppose a Horizontal Federated learning method is to be used. In that case, the machine learning model will run on each of these samples (person) individually, allowing them to train the local model and finally exchange and aggregate model weights to generate a global model.

\item \textbf{Applications}
The FedAvg algorithm proposed in \cite{shi2020communicationefficient} is the best example of a typical horizontal federated learning setup. Although horizontal federated learning protects user privacy and aims to control the communication cost, it poses challenges compared to distributed learning. Keeping in mind the protocols of federated learning and applications of neural networks, a new algorithm titled “Federated neural network” was proposed for single and multi-objective search purposes \cite{Zhu2021}. In comparison to Secure Multiparty Computation and Homomorphic Encryption, this study also analyzes how Differential Privacy may be employed best in horizontal federated learning.

EHRs have proven to be an aid for the development of scientific evidence for improving the quality of healthcare systems and they are also used as a data resource for laboratory-based public health surveillance. Techniques have been created for computing statistics on distributed databases without disclosing any personal details other than the statistical results. In a distributed dataset, duplicate records can lead to incorrect statistical findings. As a result, safe deduplication is an effective preprocessing step for improving the precision of statistical analysis of a distributed dataset. A stable protocol using a deterministic record linking algorithm for deduplication of horizontally partitioned datasets \cite{Yigzaw2017}. They introduced a new robust and scalable protocol for privacy-preserving deduplication of a horizontally partitioned dataset based on Bloom filters. According to the results of the experiments, within 45 seconds, one million virtual documents were deduplicated across 20 data custodians. The protocol is proven to be more effective and flexible than previous protocols for the same problem.

\begin{figure}[h]
  \centering
  \includegraphics[width=14cm]{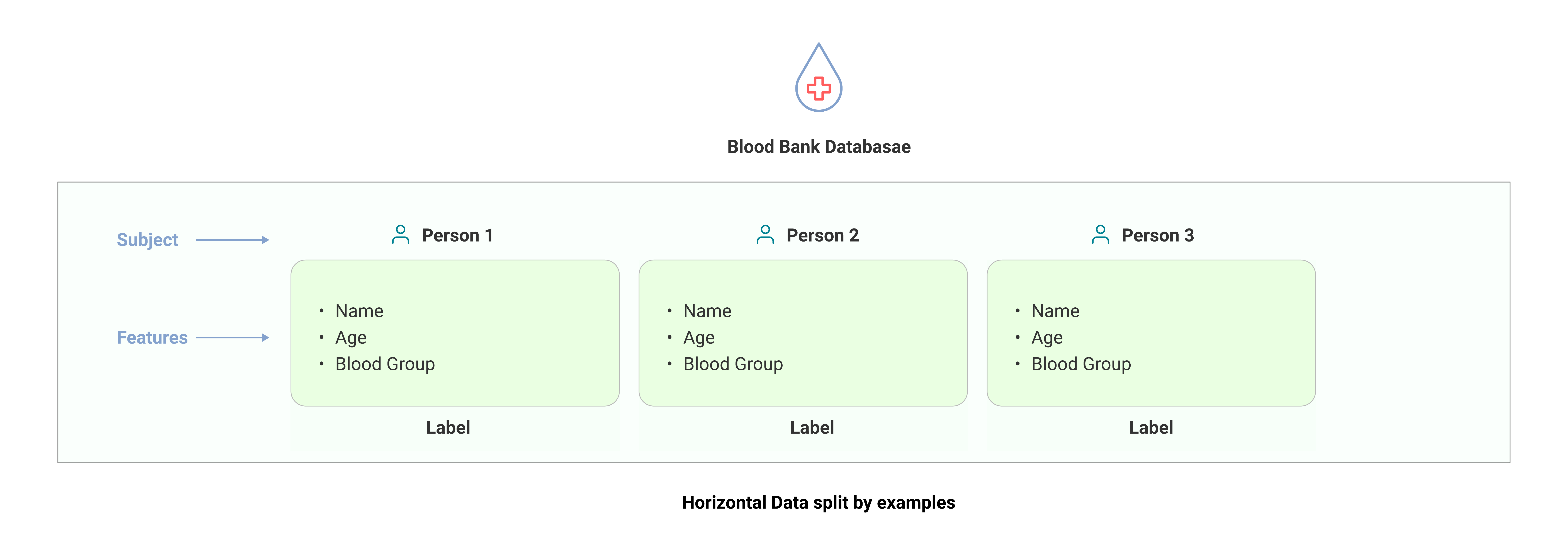}
  \caption{Horizontal Data split by examples}
  \Description{A woman and a girl in white dresses sit in an open car.}
\end{figure}

\end{itemize}
\noindent 
\item \textbf{Vertical Federated Learning}

\begin{itemize}
    \item \textbf{Concept}
    Vertical federated learning can also be called feature-based federated learning. This is used when the data shared among them contains different features but similar samples. This requires a different kind of training architecture when compared to horizontal federated learning. It may or may not involve a central server or a third neutral party. Vertical FL (VFL) applies to collective situations in which individual parties do not have access to the whole collection of features and labels and thus are unable to train a model locally using their datasets. Parties' datasets, in particular, must be synchronized to construct the full function vector without revealing their respective training data, and model training must be performed in a privacy-preserving manner. \cite{Xu2021} An example of vertical data split by examples is shown in figure 4.Assuming a blood bank laboratory in a hospital has data from the same group of patients. As seen in figure 4., the blood bank database stores the details of the patient groups under Name, Age, and Blood group. At the same time, the hospital database stores the details of the same patient groups under Name, Age, Date of Birth, Blood group, and Medical History.
    If a Vertical Federated learning method is to be used, the first dataset for the machine learning model would be the data sample from Person 1 of the Blood Bank Database and Hospital Database. Furthermore, the second dataset will be data samples from Person 2 of the Blood Bank database and Hospital Database, respectively.

\item \textbf{Applications}
FedAI uses horizontal federated learning for improving an anti-money laundering model and the use of vertical federated learning to obtain a better risk management model. Homomorphic Encryption is the privacy-preserving technique that is usually adopted when the data is distributed vertically.  Recently, a two-party design has been proposed by eliminating the trusted coordinator, which considerably decreases the system's complexity \cite{Yang2019}.

In the healthcare sector, information sharing is becoming increasingly relevant. First and foremost for medical reasons, such as the sharing of treatment records between healthcare providers, but also for secondary purposes, such as the implementation of value-based healthcare and processes. While these requirements allow information to be transferred, they also pose concerns regarding maintainability and possession, as well as protection and privacy. Provenance and permission become more complicated as data is shared by various health care providers. Sending programs containing queries and algorithms to the data source is one of the alternatives to data transfer. To address these problems, the authors in \cite{Soest2018} provide an architecture to enable algorithm conversion and execution, and use this infrastructure in a proof-of-concept setup. The proof-of-concept (PoC) focuses on processing data that has been vertically partitioned from two institutes. In a population cohort analysis, the PoC is used as a baseline to look at the causes of diabetes initiation and development, including social and environmental conditions.

Since full collections of labels and functions are not controlled by one person, privacy-preserving vertical FL is difficult. Existing vertical FL approaches necessitate multiple peer-to-peer interactions among parties, resulting in longer training periods, and are limited to (roughly) linear models and only two parties. To bridge this void, the authors of \cite{Xu2021} suggest FedV, a framework for safe gradient computing in vertical settings for a variety of commonly used machine learning models. Linear simulations, logistic regression, and support vectors are examples of these types of models. FedV uses functional encryption mechanisms to eliminate the need for peer-to-peer correspondence between parties. FedV can attain shorter preparation times as a result of this. It's also successful for wider and ever-changing groups of parties.

\begin{figure}[h]
  \centering
  \includegraphics[width=14cm]{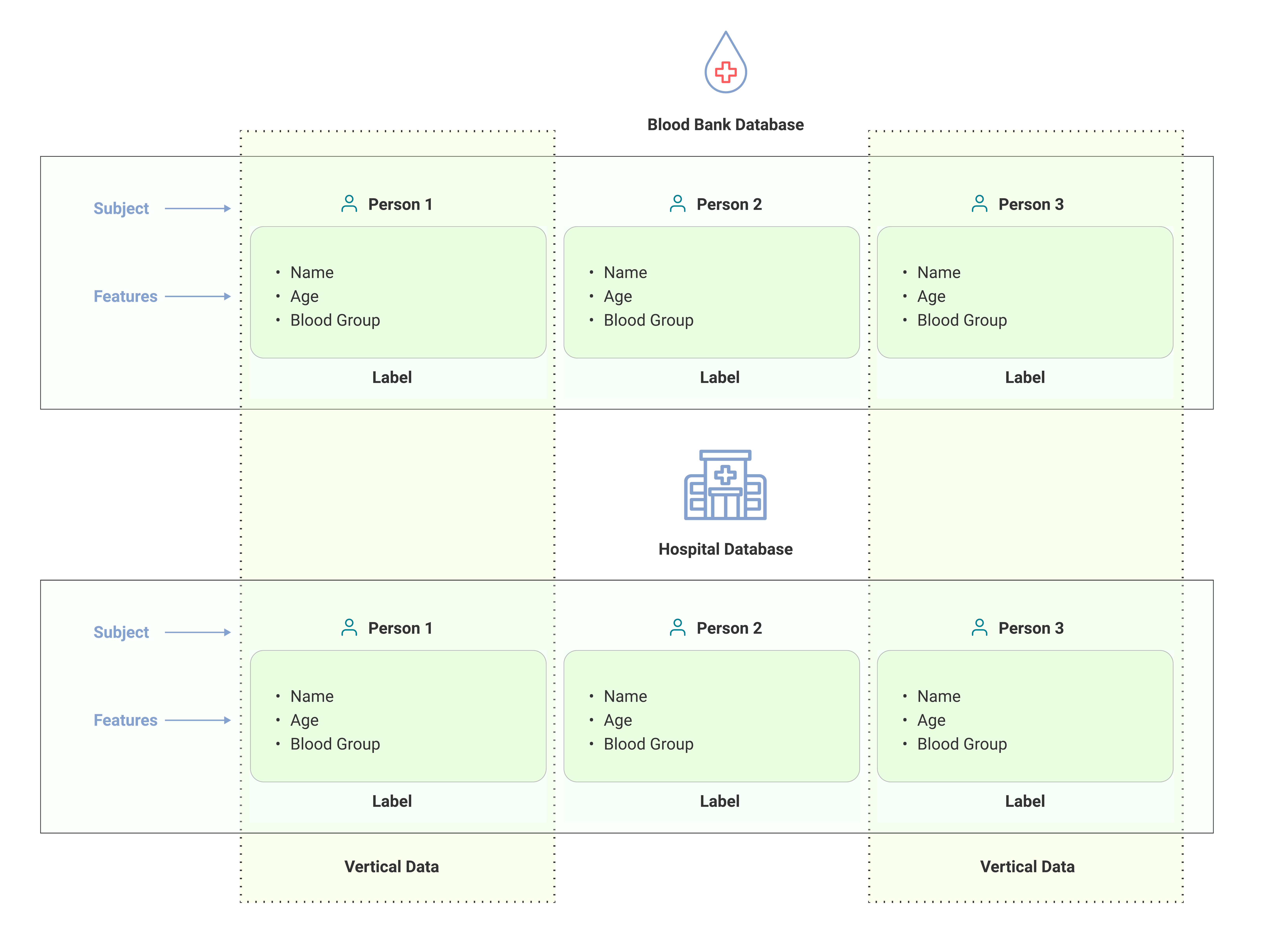}
  \caption{Vertical Data split by examples}
  \Description{A woman and a girl in white dresses sit in an open car.}
\end{figure}

\end{itemize}
\item \textbf{Hybrid/Transfer Federated Learning}

\begin{itemize}
    \item \textbf{Concept}
    Federated transfer learning \cite{10.1145/3298981} was introduced to tackle the challenge of integrating the scattered data and improve statistical modeling while performing data federation. It is also studied that this framework requires minimal modifications to the model structure and the result produced are as efficient as the non-privacy preserving transfer learning. Federated transfer learning does not depend on any requirement such as common feature space or common sample space and it supports transfer learning in providing solutions for the entire sample and feature space while data federation is taking place. 

Federated transfer learning is a term that recognizes difficult situations in which data parties only have partial overlap in the user or function space and uses current transfer learning strategies to collaboratively construct models. The current formulation is only good for two clients \cite{kairouz2021advances}. An illustration of hybrid/transfer federated learning is shown in figure 5.




\item \textbf{Applications}
FedHealth is one such algorithm that uses the concept of federated transfer learning in smart wearable healthcare devices \cite{Ju2020}. The algorithm performs data aggregation through federated learning and builds personalized models by transfer learning without compromising on the privacy and security of the model and data.

The shortage of massive datasets has hampered the progress of deep learning (DL) approaches in the field of Brain-Computer Interfaces (BCI) for the classification of electroencephalographic (EEG) recordings\cite{Ju2020}. The ability to create a large EEG BCI dataset by combining several small ones for jointly training machine learning models is limited due to privacy issues associated with EEG signals. Addressing this issue, a novel privacy-preserving deep learning architecture centered on the federated learning system, for EEG classification called Federated Transfer Learning (FTL) \cite{Ju2020}.As a result, in a subject-adaptive study, the FTL method had a 2\% better classification accuracy.

Although current federated learning systems (FLSs) primarily focus on one type of partition, data partitioning among parties in many other applications can be a combination of horizontal and vertical partitioning. As an example, consider a cancer detection method. A consortium of hospitals needs to build a FLS for cancer diagnosis, but each hospital has separate patients and medical test outcomes. In such cases, transfer learning could be an option. \cite{shi2020communicationefficient} suggest a stable federated transfer learning method that can learn a representation of a party's features using similar instances. This framework involves only minor changes to the original model layout and achieves the same degree of consistency as non-privacy-preserving transfer learning. It is adaptable and can be used for a variety of stable multi-party machine learning activities.

\end{itemize}

\begin{figure}[h]
  \centering
  \includegraphics[width=12cm]{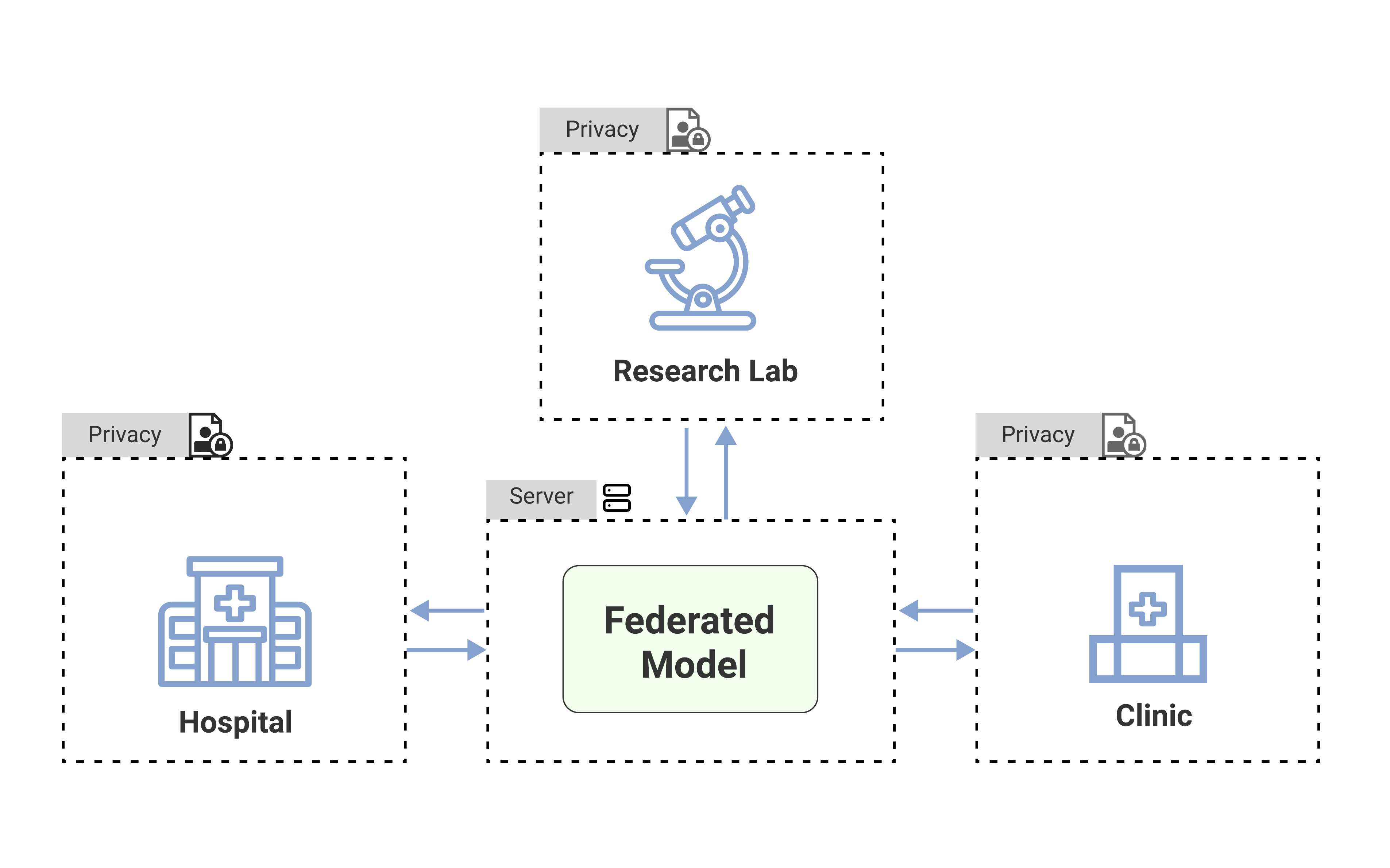}
  \caption{Transfer learning in federated learning}
  \Description{A woman and a girl in white dresses sit in an open car.}
\end{figure}

\end{itemize}

\noindent
\subsubsection{Data Pre-processing and Feature Engineering}

Data Preprocessing is an essential and challenging task in the FL training process due to the sensitive and distributed nature of the data. Moreover, it takes a long time to adapt centralized preprocessing approaches to federated data. Data preprocessing involves data cleaning, missing values imputation, and other steps to maintain consistency between distributed datasets of different clients without knowing the underlying distribution. 

A few scenarios require domain knowledge to clean, structure the data, and extract data features (properties, characteristics, attributes) from raw data. \cite{Ju2020} used medical knowledge and AI techniques to clean and normalize the raw patient data collected from hospitals. The data from the hospital includes outpatient/in-patient prescriptions, outpatient/in-patient EHRs, and other metadata.

\noindent
\subsubsection{Privacy Preservation in FL}
Federated learning involves the participation of thousands or millions of devices \cite{Yang2018} such as phones, cars, medical institutions, etc. The federated learning environment, where the model is learned locally without disclosing to any clients the input data or the output of the model, avoids direct leakage when training or using the model assuring the client’s (health systems from medical institutions in this case) dataset is kept as private as possible. This is demonstrated by a model developed using patient Electronic Health Records (EHR) and a consumer-based application such as screening atrial fibrillation with electrocardiograms obtained by smartwatches \cite{Perez2019}. Sensitive patient data is kept in local institutions or with individual consumers rather than being exposed to the federated model learning process, ensuring the patient's privacy. The ‘No peek’ rule \cite{Vepakomma2018} refers to techniques of distributed deep learning models that do not look at raw data once it leaves the clients. A good example explaining this would be, hospitals or healthcare systems at institutions that are not allowed to share data with for-profit entities due to trust issues. Such institutions are also restricted from sharing it with outside entities due to the consent of patients and various regulations like HIPAA \cite{Vepakomma2018} Listed below are few methods of privacy-preserving in FL.

Secure Privacy-preserving learning on medical data mainly involves protecting the collection of personal data used in ML model training and inference in such a way that it does not reveal any additional information about the subjects. Cryptographic techniques and differential privacy techniques are the most extensively utilized methods for privacy-preserving ML \cite{AlMohammad2019}
\begin{itemize}

\item \textbf{Cryptographic methods}
are used when encrypted data is required during testing and training phases. The widely used method under this is Homomorphic encryption and Secret sharing. Homomorphic encryption involves the encryption of data using ciphertext and public keys. It enables computations like addition and multiplication, which is a base for other complex functions, on encrypted data.

    \item \textbf{Secret Sharing}
    The technique of transmitting secrets to various parties is known as secret sharing while keeping a "share" of the secret. Only when all individual shares are merged will the secret be reconstructed. The secret is reconstructed in some configurations that won't need all shares to be merged. A privacy-preserving system for emotion detection is introduced in \cite{Hossain2019}. The authors used a multi-secret sharing method to transfer audio-visual data gathered from users to the cloud using edge devices where a CNN and sparse auto-encoder were applied for the extraction of features. The vector machine (SVM) was used for emotion and support identification.
    
    \item \textbf{Differential privacy}
    Differential Privacy (DP) is another technique to protect the privacy of individual data which has been used in areas that use algorithms like boosting \cite{Dwork2010}, principal component analysis \cite{Chaudhuri2013}, and support vector machines. Differential privacy involves the addition of noise to model updates as they communicate with servers and clients\cite{Dwork2006} . Differential privacy, on the other hand, is known to shield users from data leakage to a limited level and may lower prediction accuracy performance \cite{Cheng2019}. Hence many researchers combine DP with the Secured Multiparty computations technique to decrease the growth of noise injection and data poisoning attacks \cite{Alfeld2016}. While there is an exchange in convergence rates at the participant level - Differential Privacy (DP) may not help much in the protection of users and also may affect the accuracy of the model. Participant level DP is good when a large pool of devices are participating in the process. The ability of participant-level DP for a small pool of participants is yet to be improved and focused on working for a better convergence at this small level.
    
    \item \textbf{SMPC - Secure MultiParty Computation}
    Secured MultiParty Computation approach provides clear security guarantees even though it is in general a less efficient approach. Privacy-Preserving Record Linkage (PPRL) \cite{Laud2018} is one such example where secured multi-party computation is being used. SMC was also introduced to Quantitative Structure-Activity Relationship (QSAR) and drug-target Interaction prediction for drug discovery \cite{Laud2018}.
SMC applies the concept of three distinct responsibilities for parties. There are input parties who contribute data to the calculation that must be protected. Computation parties carry out the privacy-preserving computation on the data given. Apart from what is revealed through the architecture of the application, the computation parties should not learn anything new from the execution. Finally, there are result parties who are given the computation's results.

\end{itemize}

\begin{longtblr}[
caption = {Summary of existing studies on FL in Healthcare since 2015},
  label = {Tab:dcnnarchitectures},
                ]
                {
    rowhead=1,
    hline{1,2,Z} = 1pt, hline{3-Y},
    colsep = 3pt,
    colspec = {@{} X[j,h] Q[c,m] Q[c,m] Q[c,m] X[j,h] @{}},
    rows = {font=\footnotesize},
    row{1} = {font=\footnotesize\bfseries},
                }
Title   & Ref   & Year  & Category  & Content       \\
%

A Federated Network for Translational Cancer Research Using Clinical Data and Biospecimens
& \cite{Jacobson2015}
& 2015
& {Learning \\ Systems}
& This report describes a fully functional federated data and biospecimen sharing network for cross-institutional cancer research collaboration \\

Privacy-preserving GWAS analysis on federated genomic datasets
& \cite{Constable2015}
& 2015 & Framework
& On federated genomic datasets, this research proposes a privacy-preserving GWAS methodology \\

Privacy-Preserving Integration of Medical Data
& \cite{Miyaji2017}
& 2017 
& Protocol
& This work presents a safe and privacy-preserving method for searching and integrating health care data from diverse sources \\

LoAdaBoost: loss-based AdaBoost federated machine learning with reduced computational complexity on IID and non-IID intensive care data
& \cite{Huang2020}
& 2018
& {Learning \\ Systems}
& LoAdaBoost, a methodology for increasing the efficiency of federated machine learning, was suggested in this research, and the algorithm was evaluated using data from intensive care units in hospitals \\

Federated learning of predictive models from federated electronic health records
& \cite{Brisimi2018}
& 2018 
& Framework
& A novel FL framework is presented that can train predictive models through peer-to-peer cooperation instead of exchanging raw EHR data \\

FADL:Federated-Autonomous Deep Learning for Distributed Electronic Health Record
& \cite{liu2018fadlfederatedautonomous}
& 2018 
& {Learning \\ Systems}
& By presenting a novel approach called Federated-Autonomous Deep Learning, this study illustrates the efficacy of FL by using ICU data from 58 different hospitals to predict patient mortality can be trained quickly without transferring health data out of their silos under FL environment (FADL) \\

Patient Clustering Improves Efficiency of Federated Machine Learning to predict mortality and hospital stay time using distributed Electronic Medical Records
& \cite{Huang2019}
& 2019
& {FL in \\ Biomedical}

& The community-based federated machine learning (CBFL) technique is described in this research, and it is tested on non-IID ICU EMRs. \\

FedHealth: A Federated Transfer Learning Framework for Wearable Healthcare
& \cite{Chen2020}
& 2019
&  {FL in \\ Healthcare IoT}
& To address data privacy concerns, this paper proposes a federated transfer learning system for wearable healthcare \\

Communication-Efficient Federated Deep Learning with Asynchronous Model Update and Temporally Weighted Aggregation
& \cite{Chen2020}
& 2019
& {Learning \\ Systems}
& This paper presents a synchronous learning strategy for FL clients \\

Federated deep learning for detecting COVID-19 lung abnormalities in CT: a privacy-preserving multinational validation study
& \cite{Dou2021}
& 2019
& Diagnosis
& With external validation on patients from a global cohort, this report reveals the efficiency of an FL system for identifying COVID-19 associated CT anomalies \\

Federated Learning for Healthcare Informatics
& \cite{XuJ2021}
& 2019
& Survey
& This survey study provides an overview of federated learning systems, focusing on biomedical applications \\

Federated electronic health records research technology to support clinical trial protocol optimization: Evidence from EHR4CR and the InSite platform
& \cite{Claerhout2019}
& 2019
& {FL in\\ EHR data}
& This paper determines if inclusion/exclusion (I/E) criteria of clinical trial protocols can be represented as structured queries along with those executed using a secure federated research platform (InSite) on hospital electronic health records (EHR) \\

Predicting Adverse Drug Reactions on Distributed Health Data using Federated Learning
& \cite{Choudhury2019}
& 2019
& Framework
& To increase the global model's predictive power, this research proposes two unique approaches to local model aggregation \\

Privacy-preserving Federated Brain Tumour Segmentation
& \cite{Li2019}
& 2019
& {FL in \\ Biomedical}
& Adopting the BraTS dataset for brain tumor segmentation, this research investigates the possibility of using differential-privacy approaches to secure patient data in a federated learning context \\

Multi-site fMRI Analysis Using Privacy-preserving Federated Learning and Domain Adaptation: ABIDE Results
& \cite{Li2020}
& 2020
& {Medical Image \\  Analysis}
& This work proposes a privacy-preserving multi-site fMRI classification that ensures that private information cannot be retrieved from model gradients or weights \\

Stochastic Channel-Based Federated Learning With Neural Network Pruning for Medical Data Privacy Preservation: Model Development and Experimental Validation
& \cite{Shao2020}
& 2020
& {Learning \\ Systems}
& For the study of distributed medical data, this research proposes a privacy-preserving approach called stochastic channel-based federated learning (SCBFL) \\

Federated learning in medicine: facilitating multi-institutional collaborations without sharing patient data
& \cite{Sheller2020}
& 2020
& Medicine
& This research shows that utilizing data from ten universities, the models achieve 99 percent model quality and discuss the impact of data distribution across participating institutions \\

FedHome: Cloud-Edge based Personalized Federated Learning for In-Home Health Monitoring
& \cite{WuQ2020}
& 2020
& {FL in \\ Healthcare IoT}
& FedHome, a cloud-edge-based federated learning architecture for in-home health monitoring, is proposed in this research \\

The future of digital health with federated learning
& \cite{Rieke2020}
& 2020
& Survey
& This survey report looks at how FL could help with the future of digital health, as well as the obstacles \\

Federated Learning on Clinical Benchmark Data: Performance Assessment
& \cite{Lee2020}
& 2020
& Benchmark
& The research uses three benchmark datasets, including a clinical benchmark dataset, to assess the reliability and performance of FL \\

FedMed: A Federated Learning Framework for Language Modeling
& \cite{Wu2020}
& 2020
& Framework
& To address model aggregation and communication costs in the FL environment, this study provides a unique Federated Mediation (FedMed) framework with adaptive aggregation, mediation incentive scheme, and topK method \\

Federated Learning for Breast Density Classification: A Real-World Implementation
& \cite{Roth2020}
& 2020
&  {Medical Image \\  Analysis}
& This article demonstrates the efficacy of FL by training a model for breast density categorization based on Breast Imaging, Reporting, and Data systems utilizing data from seven clinical institutions across the world (BI-RADS) \\

Federated Transfer Learning for EEG Signal Classification
& \cite{Ju_2020}
& 2020
& {Learning \\ Systems}
& This work proposes a unique privacy-preserving DL architecture called federated transfer learning that uses the FL in EEG classification (FTL) \\

COVID-19 detection using federated machine learning
& \cite{abdul2021covid}
& 2021
& Diagnosis
& To determine which parameters impact model prediction accuracy and loss, this study employed a descriptive dataset and chest x-ray (CXR) images from COVID-19 patients in an FL context \\

Implementing Vertical Federated Learning Using Autoencoders: Practical Application, Generalizability, and Utility Study
& \cite{Cha2021}
& 2021
& {Learning \\ Systems}
& Without revealing the raw data, this research shows that FL on vertically partitioned data may perform equivalent to centralized models \\

Federated Learning Meets Human Emotions: A Decentralized Framework for Human–Computer Interaction for IoT Applications
& \cite{Chhikara2021}
& 2021
& {FL in \\ Biomedical}
& This article combines facial expression and voice inputs to construct an emotion monitoring \& analysis system using FL \\

FeARH: Federated machine learning with anonymous random hybridization on electronic medical records
& \cite{Cui2021} 
& 2021
& {Learning \\ Systems}
& This research study suggests a novel FL method to deal with untrustworthy conditions \\

Federated Learning for Thyroid Ultrasound Image Analysis to Protect Personal Information: Validation Study in a Real Health Care Environment
& \cite{Lee2021}
& 2021
& Diagnosis
& The purpose of this research is to see if FL's performance is equivalent to that of traditional deep learning \\

Learning From Others Without Sacrificing Privacy: Simulation Comparing Centralized and Federated Machine Learning on Mobile Health Data
& \cite{Liu2021}
& 2021
& {FL in \\ Healthcare IoT}
& The research explores FL use cases in a mHealth environment and uses a mHealth data set to simulate federated learning \\

Cloud-Based Federated Learning Implementation Across Medical Centers
& \cite{Rajendran2021}
& 2021
& {FL in\\ EHR data}
& This research mimics an FL environment in order to investigate multiple federated learning implementations and apply FL algorithms to data from two academic medical facilities' electronic health records \\

Federated learning improves site performance in multicenter deep learning without data sharing
& \cite{Sarma2021}
& 2021
& {FL in\\ EHR data}
& This study demonstrates how to provide multi-institutional training in a FL environment without centralization \\

A Resource-Constrained and Privacy-Preserving Edge-Computing-Enabled Clinical Decision System: A Federated Reinforcement Learning Approach
& \cite{Xue2021}
& 2021
& {FL in\\ EHR data}
& This article combines mobile-edge computing (MEC) with software-defined networking to make use of the processing and storage capabilities available among edge nodes (ENs) (i.e., MEC servers) in the Fl environment \\

Variation-Aware Federated Learning with Multi-Source Decentralized Medical Image Data
& \cite{Yan2021}
& 2021
& {Learning \\ Systems}
& Variation-aware federated learning (VAFL) is a methodology proposed in this research for minimising client variations by transforming all clients' images into a shared image space \\

Federated Learning in a Medical Context: A Systematic Literature Review
& \cite{Pfitzner2021}
& 2021
& Survey
& This survey article examines federated learning and its relevance to sensitive healthcare data \\

Federated Learning for Smart Healthcare: A Survey
& \cite{Nguyen2021}
& 2021
& Survey
& The application of FL in smart healthcare and IoT devices are reviewed and surveyed in this survey report \\

\end{longtblr}

\noindent
\subsection{\textbf{Machine Learning Pipeline}}
The life cycle of a federated learning system consists of 8 primary steps which include task initialisation, selection, configuration, model training, client server communication, scheduling and optimisation, versioning testing deployment and termination. The steps are illustrated in the diagram shown below. (fig 6)

\begin{figure}[h]
  \centering
  \includegraphics[width=\linewidth]{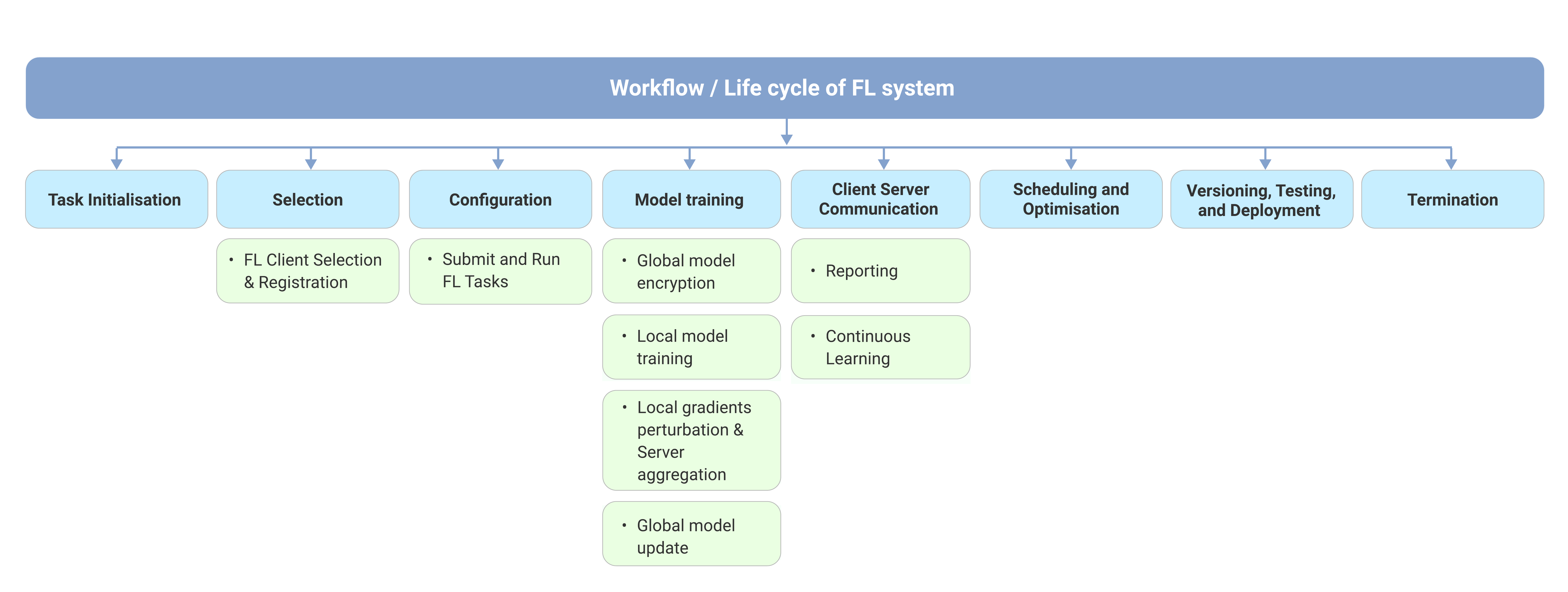}
  \caption{Machine Learning Pipeline}
  
\end{figure}

In FL settings, three major machine learning steps include model selection, model training, and hyperparameter tuning.

\subsubsection{Model Selection}
The model selection stage includes the process of selecting the optimal model for federated data. Model selection depends on the type of task as well as size, quality, and type of the federated data. For example, if the problem is classification on diagnostic images, A convolution neural network is the best choice for this task. There have been several attempts in recent years to propose and create new models for federated settings. However, most FL tasks consider state-of-the-art, widely used models in their setting. Neural networks (NN) are the most popular and state-of-the-art in many machine learning tasks in the FL setting. For example, \cite{hard2019federated} uses a variant of LSTM called Coupled Input and Forget Gate  (CIFG) to predict the next word in the mobile keyboard. \cite{chen2021fedhealth} uses a specific Convolutional Neural Network (CNN) network with a transfer learning method for the classification task. A tree-based FLS has shown impressive performance on classification and regression tasks. \cite{li2021privacypreserving} used the FLSs for Gradient boosting decision trees (GBDTs) on horizontally and vertically federated data, respectively.  Most Federated Learning frameworks use stochastic gradient descent methods to optimize machine learning models such as neural networks and logistic regression. Besides Neural Networks and tree-based networks, linear models (e.g., linear regression, support vector machines) are standard and easy-to-use methods.

\noindent
\subsubsection{Model Training}

Following the selection of a model, the next stage in the ML pipeline is model training. Different variants of the model and a combination of optimizing hyperparameters are used to decide the final model with good accuracy. This process \cite{DBLP:journals/corr/abs-2007-00914} involves the following steps 
\begin{itemize}
    \item Training the local models on their local training dataset

\item Sharing of the local parameters to the server

\item Aggregation of local models’ parameters on the server using the aggregation operator and

\item Updating the local models with the aggregated global model
\item Repeat the loop

\end{itemize}

\subsubsection{Model Parameters}
The parameters are learned from federated data, and hyper-parameters are utilized to fine-tune the output for the best match. The hyper-parameters such as learning rate, number of training epochs, mini-batch size, and optimizer have to be tuned based on the constraints of the ML application (e.g., available computing power, available memory, bandwidth). \cite{wang2019federated} demonstrated how local data could be used to fine-tune federated models. They provided methods for determining the appropriate hyper-parameters for fine-tuning and proved that it enhances models' next word prediction in mobile keyboards. \cite{yu2020salvaging} offered numerous variations of the fine-tuning strategy to improve the local adaptation. 
The network architecture is designed manually, which takes a significant amount of effort and experience in that field. Neural Architecture Search (NAS) is an algorithmic-based method to search the neural network design utilizing optimization algorithms. In recent years there has been progress in federated neural architecture search. It aims to optimize the design of models in the federated learning environment. The offline federated NAS framework proposed by \cite{zhu2020federated} uses a multi-objective evolutionary algorithm to design the optimal network.

\section{\textbf{Challenges \& Issues}}

Some various challenges and issues can be found at each step in the implementation of a federated learning system. In this paper, we classify the challenges and issues into three major subcategories. The first subcategory involves all issues related to the privacy of the model and the attacks that happen on the system. The second subcategory consists of papers that describe the issues related to data such as limited data, data bias, or data poisoning. 

The kinds of communication that can take place in a federated learning setup are between a server and an edge device, between two edge devices, and or between more than one server. The main challenge here is to answer some of the following questions: how effective can this communication be without having to lose the participating devices? How can the number of communication rounds be reduced to avoid large usage of battery or net on edge devices? And, how can all of this be done without affecting the model and its result metrics? This subsection brings down papers addressing these major issues. Hence, the third subsection speaks about communication challenges. An illustration of this section is presented in figure 7.

\begin{figure}[h]
  \centering
  \includegraphics[width=\linewidth]{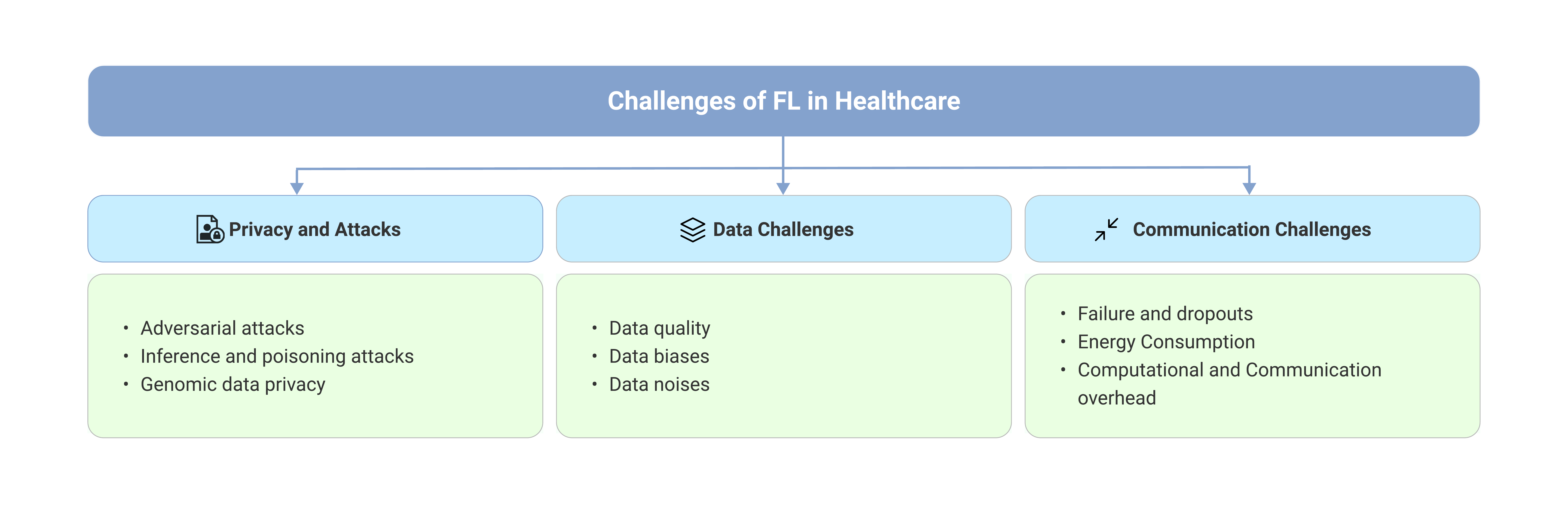} 
  \caption{Challenges of Federated Learning in healthcare}
  \Description{A woman and a girl in white dresses sit in an open car.}
\end{figure}

\subsection{\textbf{Privacy and attacks}}
\begin{itemize}
    \item \textbf{Adversarial Attacks}
    Improper or inadequate learning refers to cases where improper hyperparameters are learned in the ML/DL model, e.g., learning rate, epochs, batch size. In a predictive healthcare setting, machine learning models are created using previous patient data and then evaluated on new patients, raising concerns about the validity of the predictions.This is because of distribution shifts.And such differences can be exploited for generating adversarial examples \cite{papernot2016science} which is now a huge concern.  In addition, ML/DL versions are strictly susceptible to various risks to the protection and privacy, such as adversarial attacks \cite{szegedy2014intriguing,biggio2013poisoning}
    
    \item \textbf{Inference and poisoning attacks}
    Federated learning systems are yet to be aware of the future federated learning algorithm design on privacy preservation. \cite{lyu2020threats} This survey on threats to federated learning provides a concise introduction to the topic of federated learning and the two major federated learning methods are exposed to, inference attacks and poisoning attacks. It also summarizes the kinds of threat models a system can be prone to and highlights the key techniques as well as fundamental assumptions adopted by various attackers.
    \item \textbf{Evasion Attack}
    Evasion/exploratory attacks are among the most common adversarial attacks and are carried out during the inference time. This setting does not imply any influence over the training data. In this attack, an adversary attempts to evade the deployed model by feeding malicious data samples or collecting evidence about the model features during the inference phase. The amount of knowledge available to the adversary about the model determines the efficiency of such attacks. Various methods have been presented to make the FL models more robust against evasion attacks.

\item \textbf{Inadequate model training}

Machine learning models while training may involve flaws. The model training flaws include improper or inadequate training, violations of privacy, model poisoning, or theft. Improper or inadequate learning refers to cases where improper parameters are learned in the ML/DL model, e.g., learning rate, epochs, batch size. The deployment of the model involving ML/DL techniques mainly revolves around human-centric decisions. And hence, considering fairness, accountability while ensuring the robustness of the system is crucial at this stage. The deployment stage can involve Evasion attacks, System Disruption, Network Issues, etc.

\item \textbf{Gradient Inversion Attack}

This attack strategy implies that it is possible to recover and reconstruct input data from the gradient knowledge of trained and untrained model parameters. \cite{geiping2020inverting} investigates the impact of design and settings on the difficulty of reconstructing an input image from grading data and demonstrates that any input to a fully connected layer could be reconstructed analytically without regard to the rest of the model design.

\item \textbf{Privacy in Genomic Data}

The human genome is a complete set of genetic information of a human living organism composed of 4 different bases (A, T, G, C) and can provide a treasure of highly sensitive and personal information of an individual. With the innovation in the next-generation technologies, a whole new genome complex can be determined of an organism. The use of genome sequencing is used for purposes like personalized genomic medicine, disease diagnosis, and preventive treatment. \cite{AKGUN2015103} discusses the various privacy issues in genomic data processing such as querying on genomic data and carrying alignment processes on commercial public clouds in a privacy-preserving and effective manner. The paper also concludes that despite the innovations and study in medicines and health science, the use of sequencing technology still lacks privacy preservation of genomic data leading to a lot of genomic data leakage.
\end{itemize}

\subsection{\textbf{Challenges related to Data}}
\begin{itemize}
    \item \textbf{Noises in different kinds of data}
    The healthcare industries face a huge threat when it comes to data collection. Large amounts of clinical data are collected in the form of EHRs, medical images, radiology reports, etc.. which requires a lot of human effort and is time-consuming. Multishot MRI, one of the widely used imaging modalities used to acquire high-resolution medical images can involve Instrumental and Environmental Noise due to some undesirable artifacts in the resulting image \cite{qayyum2020secure}
A variety of health data including patient data from multi-omic approaches, as well as clinical, behavioral, environmental, and drug data can be analyzed using the AI technologies being developed at the present. \cite{WangAI2019} mentions five major types of data used in AI for health such as multi-omics data, clinical data, behavioral/wellness data, environmental data, as well as research and development data. Each data type has its challenges, implications, and future directions.

\item \textbf{Improper annotation of data}

Healthcare datasets involve annotation of data samples which is a crucial step in the process of healthcare applications. Hence, they should be performed by legally keeping privacy concerns in mind and with proper guidelines. The inability to perform labeling rightly leads to improper annotations and many efficiency challenges like imbalanced dataset, class imbalance, and Bias and data sparsity \cite{qayyum2020secure}. 

\item \textbf{Data biases}
Data biases \cite{kairouz2021advances} act as the main driver of unfairness in machine learning models. It can result in high risks when used in federated learning systems. Moreover, biases can end up affecting areas like training data, cognitive sampling, reporting, and confirmation biases.
Artificial intelligence (AI) models often need a vast volume of high-quality training data, which contrasts sharply with the existing drug development pipelines' limited and skewed data. And, this decentralized machine learning paradigm has a high scope in contributing to the improvement of AI-Based drug discovery \cite{Xiong2020FacingSA}.  The superiority of the federated learning mechanism is demonstrated on pooled datasets with 7 aqueous solubility datasets including high and low biases. The authors also address the small data and biased data dilemma in drug discovery and prove the promising role of federated learning.

\item \textbf{False positives and false negatives}
The Healthcare industry deals with a large amount of patient data and this involves high chances of containing missing observations or variables. However, ignoring these values during analysis by knowing their relationships with already observed and unobserved data is the simplest way to avoid them. On the other hand, using these missing observations leads to well-known problems like false positives and false negatives and thus there is a need for complete and compact healthcare data. Issues like false positives and false negatives not only are caused due to the use of missing values but also due to incomplete training or inefficient training of the model. The root cause behind inefficient training is the incomplete data being fed for inference. ML-powered healthcare demands cautious applications of analytical methods \cite{Pollard2019}. And that is why along with quantity, quality of data also plays a major role. 
\item \textbf{Data Quality}
\cite{carlsson2020privacy} A survey paper on applicable machine learning algorithms in a federated environment mentions the importance of Independent and Identically Distributed (IID) Data points and their contribution to the factor of lowering the chances of class imbalance. Many times the local data collected can lack in size and major data points which may obstruct representing the underlying structure of a federated learning system. Using such a small-sized imbalance dataset has high chances of hindering the globe model and its accuracy. 
\cite{Jochems2017DevelopingAV} has trained the global model on historical patient data of Radiation Therapy (RT); because RT technique is constantly evolving, such models provide little value to clinical practice today. In addition, Phase III clinical trials give high-quality evidence, but they have the drawback of taking a long time to complete. However, Including more recent patient data and updating the model with the most up-to-date practice insights might help alleviate the problem. 
According to the author of the \cite{Yigzaw2020PrivacypreservingAF}, federated learning for an institution/hospital takes several months to develop. Furthermore, obtaining data from electronic health records(EHR) remains challenging since data is typically dispersed across several databases and apps. 

\item \textbf{Data Leakage}
Federated learning provides a solution that enhances data privacy, which is a crucial concern of FL training. However, some work has proposed showing the leakage in federated learning representing a still unexplored area of research due to several factors that may result in security issues. A quantifiable method of measuring privacy would enable better choices about the minimum privacy settings required to sustain clinically acceptable results \cite{Flores2021FederatedLU}.

\item \textbf{Patient Variability}

Care varies widely between locations, with significant variation in performance across indicators. Hospital levels could explain more of the observed diversity than clinician levels.

In a study, more than 7,000 cardiovascular patients were treated with adult stem cells \cite{FernndezAvils2017GlobalPP} 
However, the data so far has revealed neutral or minor advantages. One of the most important questions to explore when thinking about designing better trials is why there is so much variation both across and within trials. Patients in the FOCUS-CCTRN trial \cite{Perin2012EffectOT} received autologous bone marrow cells to treat chronic ischemic heart disease.  Several cardiac clinical outcomes showed no improvement when compared to placebo. On the other hand, individual patient outcomes were highly varied \cite{Beachy2019ExploringSO}.

Care variability or uneven care practices has an impact on many elements of healthcare delivery. It can assist improve patient safety, reducing healthcare expenditures, and improving key performance indicators (KPIs) for both people and the healthcare system by minimizing variability of care.

\end{itemize}

\subsection{\textbf{Communication Challenges}}

\begin{itemize}
    \item \textbf{Failure or dropouts during communication}
    While distributed learning conjointly aims at training one model on multiple servers, a standard underlying assumption is that the native datasets are Identically Distributed (IID) and roughly have constant size. None of those hypotheses are created for federated learning; instead, the datasets are usually heterogeneous and their sizes could span over many orders of magnitude. Moreover, the clients concerned in federated learning could also be unreliable as they're subject to a lot of failures or drop out. Some clients' models do not get included in every FL round due to interrupted connectivity or slow internet conditions. Those devices or models are called stragglers. \cite{Flores2021FederatedLU} Since the clients usually trust less powerful communication media (i.e. Wi-fi) and powered systems (i.e. smartphones and IoT devices) compared to distributed learning where nodes are usually data centers that have powerful procedure capabilities and are connected to at least one another with quick networks.

For an Edge AI task to be accomplished there are multiple communication rounds between edge nodes. Edge nodes are usually smartphones or devices. Small access to training data is available at every edge node that is participating in the training. These nodes then perform edge training which results in high communication costs especially in case of limited bandwidth. \cite{Zhao2018} The aim of communication in every round is to compute a certain function value concerning intermediate value at edge devices. The paper points out a challenge of alleviating communication overheads under privacy and resource constraints and the need of reducing the communication rounds for training and inference. The paper also introduces communication efficient methods to achieve efficient results. Communication methods for edge AI at the algorithmic level, zeroth-order method, first-order method, second-order method, and federated optimizations have been well illustrated and explained.

\item \textbf{Computational and Communication Overhead}

Communication has become one of the primary challenges for federated learning as the wireless networks and end-user internet connections can potentially become expensive and unreliable soon. \cite{Kairouz2021} mentions how federated averaging and sparsification and/or quantization of model updates have demonstrated a significant reduction in communication cost with minimal impact on training accuracy. 
\cite{Constable2015} demonstrates that employing secure MPC experiments to do privacy-preserving federated genomic data analysis is more costly than performing the exact computation in a centralized non-encrypted environment.

\item \textbf{Energy consumption while communication}

IoT(Internet of Things) involves widespread use of mobile devices involving computing and sensing capabilities which involves a collection of data at a societal scale. Valuable data collection and maintenance backed with centralized machine learning models entail security and privacy issues leading to less participation of devices in smart city methods. While the mechanics of privacy preservation in federated learning ensures the privacy of data is preserved, mobile crowdsensing involves a huge amount of energy consumption. Federated learning involves on-device training in turn earning leading to high consumption of batteries from local devices which inturn deters users from participating in the process. For instance, this \cite{Jiang2020FederatedLI} survey paper presents the potential of federated learning along with the overview of challenges and issues faced while incorporating federated learning into smart city sensing.

\end{itemize}

\section{\textbf{Application}}

\begin{figure}[h]
  \centering
  \includegraphics[width=\linewidth]{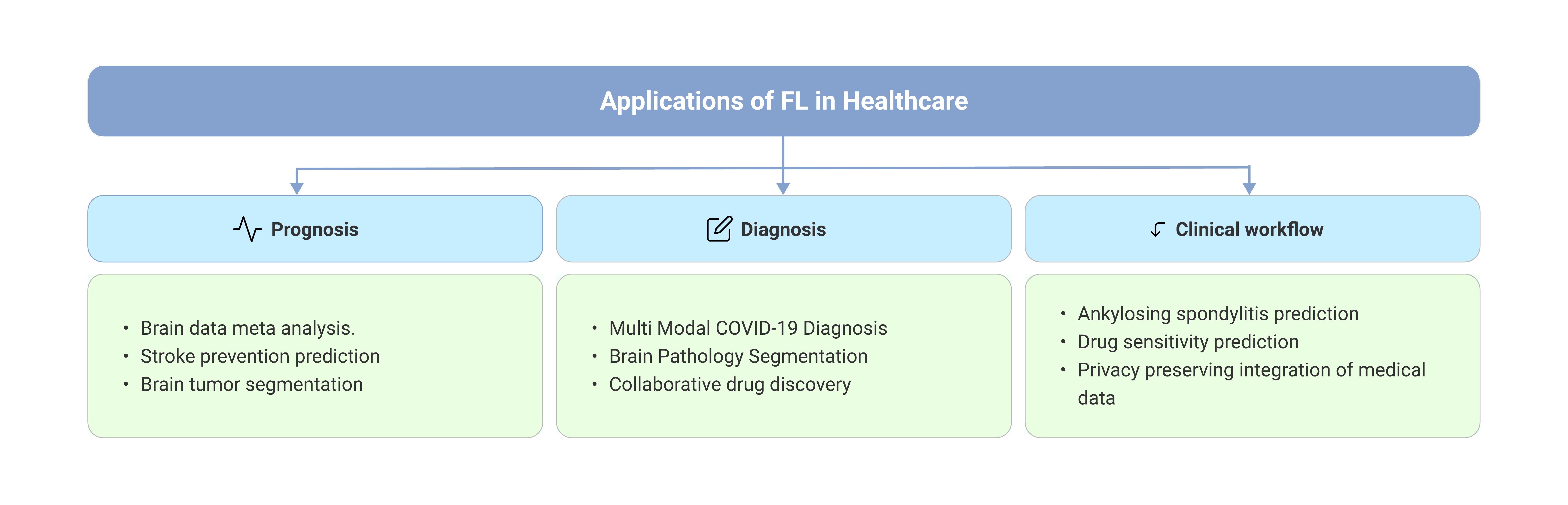}
  \caption{Applications of federated learning in healthcare}
  \Description{A woman and a girl in white dresses sit in an open car.}
\end{figure}

\subsection{\textbf{Prognosis}}
Prognosis is a diagnostic term that refers to assessing the possible or potential progression of an illness, such as whether the signs and effects will change or worsen (and how quickly) or stay constant over time. The natural course of the diagnosed illness, the individual's physical and emotional health, the current medications, and other considerations are used to make a prognosis. Listed below are healthcare-related applications helping to perform a prognosis on a particular disease or type of data.

 \textbf{Privacy preserving stroke prevention}
    An investigation into Facebook's data protection breaches in processing consumer information for uninformed usage has sparked recent data privacy issues. As a result, legislation such as the General Data Protection Regulation (GDPR) [Regulation, 2016] has been proposed to prevent organizations from sharing data without prior user consent. To fix data privacy, they use a new and efficient methodology called federated learning platform, which allows them to jointly train a machine learning model using data from multiple clients without directly exchanging data between them. Tencent and WeBank collaborated to create a privacy-preserving stroke prediction technology. Stroke prevention, as well as the risk factors associated with it, has long been a public health priority around the world. The scientists and engineers suggest a privacy-preserving scheme for predicting stroke risk and intend to use cloud servers to deploy the federated prediction model \cite{Ju2020}.
    
    \textbf{Meta analysis on brain data}
    In reality, data sharing is limited by the need to migrate vast amounts of biomedical data, as well as the administrative workload that comes with it. Researchers sought an analysis approach in meta-analysis or federated learning paradigms as a result of this situation. The Enhancing NeuroImaging Genetics by Meta-Analysis (ENIGMA) consortium is one of the best examples of this kind of research method. Brain scans of previously unimaginable amounts can be found in data banks all around the world. Due to various privacy and legal considerations, separate databases held at multiple locations cannot always be exchanged directly, restricting the use of big data in the study of brain disorders. A Federated learning platform allows them to jointly train a machine learning model using data from multiple clients without directly exchanging data between them. Addressing the issue of privacy and restrictions on sharing the data, the authors propose a federated learning system for safely accessing and meta-analyzing any biomedical data while maintaining individual privacy. They demonstrate the framework by using the ENIGMA Shape platform to provide the first implementation of federated analysis that is consistent with ENIGMA's standard pipelines \cite{Silva2019}.
    
     \textbf{Brain tumor segmentation}
    Classification of electroencephalographic (EEG) recordings and brain tumor segmentation was difficult due to the need of large datasets \cite{Ju_2020,Li2019}. This lack of large datasets has led to the success of deep learning (DL) methods in the field of Brain-Computer Interfaces (BCI). A novel privacy-preserving DL architecture has been suggested for EEG classification called federated transfer learning (FTL). A test proposed architecture's success on the PhysioNet dataset for 2-class motor imagery classification. \cite{Ju_2020} Demonstrating the possibility of implementing differential-privacy strategies to secure the patient data in a federated learning setup, a brain tumor segmentation was performed on the BraTS dataset \cite{Li2019}.
    
     \textbf{Breast density classification\cite{Roth2020}}
    For the advancement of clinically relevant models, hospitals and other academic institutes often need to partner and host centralized databases. Owing to data protection and ethical issues involved with data sharing in healthcare, this overhead will easily become a technical problem and typically necessitates a time-consuming approval process. And if these issues are overcome, data is precious, and organizations may choose not to exchange complete datasets. In a real-world collaborative environment, the use of federated learning (FL) to create a medical imaging classification model was demonstrated bringing together seven health organizations from around the world to train a model for breast density classification focused on Breast Imaging, Reporting, and Data System (BIRADS) \cite{Roth2020}.
    
    \textbf{Multi-Disease Chest X-ray classification}
    
    Deep learning approaches can yield important results in medical imaging analysis, although they involve a large volume of high-quality data. Since a client cannot have enough data to train and construct a quality model, working with multiple clients may address data insufficiency problems in deep learning but add privacy restrictions. This paper \cite{McMahan2017} proposes a federated deep learning method for multi-disease classification from chest X-rays, with pneumonia as an example. The FDL technique measures pneumonia from a chest X-ray and distinguishes between viral and bacterial pneumonia. Clients train local models with minimal private data at the edge server and send them to the central server for global aggregation without sending the chest-X-ray images to a central server \cite{Banerjee2020}.
    
     \textbf{Adverse Drug Reaction prediction}
    
    Since healthcare data is scattered, collecting a sufficiently comprehensive dataset to track rare cases entails combining data from various data silos. Analyses derived from various data sources may be contradictory or inaccurate, necessitating the use of tools to properly aggregate the information. A time lag exists between the ADR (adverse drug reaction) case, claim filing, adjudication, and claim consolidation into a database of current claims-based systems. As a result, there is an unmet need for reliable, flexible, and effective methods for forecasting ADRs using distributed health data while maintaining patient privacy. Since healthcare data is dispersed, assembling a robust dataset to document unusual events necessitates merging data from several data silos. To address the issue a federated learning-based system has been proposed, which allows health data to be shared across several platforms. Without ever transferring the raw data from their respective pages, the architecture helps one to train a global model based on each site's local data. It is the first time federated machine learning algorithms have been used to forecast ADRs using distributed electronic health data \cite{Choudhury2019}.
    
     \textbf{Predictions on SARS-COV-2 chest X Rays}
    
    Confidential data all pose realistic difficulties when using electronic health data to forecast adverse drug reactions (ADR). Another example where there is a need for data collaboration is within clinical and science communities. When adapting to rapidly changing and pervasive environmental threats, the pandemic has highlighted the importance of quickly conducting data collaborations. One recent work on an AI-based SARS-COV-2 Clinical Decision Support (CDS) algorithm is a concrete example of these forms of partnerships. Another example is during the SARS-COV-2 pandemic, 20 institutes partnered on a healthcare FL study that used vital signs, laboratory results, and chest x-rays to predict possible oxygen needs of infected patients, resulting in the “EXAM” (EMR CXR AI Model) \cite{Flores2021FederatedLU}. 
    
     \textbf{GWAS analysis on genomic datasets}
    
    The biomedical community benefits from the growing availability of genomic data for scientific studies, such as Genome-Wide Association Studies (GWAS). However, high-quality GWAS typically necessitates a large number of tests, which may exceed a single institution's ability. Concerns regarding patient safety and clinical knowledge security arise from federated genomic data analysis (as data are being exchanged across institutional boundaries). On federated genomic databases, a privacy-preserving GWAS architecture. has been proposed where computations are layered on top of stable multi-party computation (MPC) structures \cite{Constable2015}.
    
     \textbf{NeuroLOG}
    A creation of a framework OntoNeuroLOG was implemented by federating five neuroimaging data repositories in Paris, Rennes, Grenoble, and Sophia Antipolis. The creation of the framework OntoNeuroLOG and its use to confirm heterogeneous data to a standard model are the main features of this work. The project focuses on research implementations that need a multi-center, multi-disciplinary approach: 1) epilepsy (surgical treatment of drug-resistant epilepsy) and 2) neurodegenerative disorders (Alzheimer's disease) \cite{Gibaud2011}

\subsection{\textbf{Diagnosis}}

The description of the type and origin of a certain phenomenon is referred to as diagnosis. Diagnosis is used in many different fields to assess "cause and effect," with differences in the application of logic, analytics, and practice. Listed below are healthcare-related applications helping to perform a diagnosis on a particular disease or type of data such as drug discovery, COVID-19 prediction at the edge, and the use of EHRs to improve mortality prediction.

 \textbf{COVID-19 diagnosis at the edge}
    
    COVID-19 has been a major focus of research in 2020, especially after the World Health Organization (WHO) declared it a pandemic in March, with various activities focused on diagnosis, prevention, and the production of a possible vaccine. Risk identification \cite{Pal2021}, touch monitoring, false news identification, emotion analysis, and screening and diagnosis are some of the main applications of data science approaches, especially machine learning and data visualization techniques, in the international response to the COVID-19 pandemic. Despite major advancements in recent years, cloud-based healthcare systems appear to be underutilized due to their shortcomings in meeting rigorous protection, privacy, and quality of service criteria (such as low latency). The authors take advantage of edge computing's capabilities in medicine by exploring and testing the ability of intelligent analysis of clinical visual data linked to COVID19 at the edge. They also enable remote healthcare centers to benefit from a multi-modal shared learning model without having to share any knowledge about the local data's modality or the data itself. The authors suggest a CFL-based collective learning system for the role of COVID19 diagnosis with visual evidence such as X-rays, Ultrasound images, and CT Scans, based on an emerging idea of clustered federated learning (CFL) \cite{Qayyum2021}
    
    \textbf{Estimation of blood pressure}
    
    According to the World Health Organization (WHO), chronic heart disease was the leading cause of death from 2000 to 2019, accounting for 16\% of all global deaths in 2019. During this time, most deaths have been caused by heart disease. This not only has a significant impact on the lives of those involved but also on public healthcare services. Electrocardiogram (ECG) and blood pressure (BP) readings are widely used by clinicians to consider the dynamics between the healthy and dysfunctional core. These methods are also very invasive, particularly when continuous arterial blood pressure (ABP) readings are taken, and they are often quite expensive. To address the problem, the authors present a decentralized learning approach to continuous ABP calculation that is capable of large-scale real-world deployment while protecting patient privacy. This architecture, to their knowledge, is the first example of a GAN capable of continuous ABP generation from an input PPG signal and using a federated learning methodology \cite{Brophy2021}.
    
     \textbf{Data Variability in Medical Imaging}
    
    Deep learning has made rapid strides in image recognition and target detection in recent years. These advancements have also led to improvements in automating clinical processes within medical imaging, thanks to Convolutional Neural Networks (CNNs) pattern recognition abilities. Deep CNNs, for example, has paved the way for breakthroughs in retinopathy diagnosis, lung nodule identification, and brain tumor segmentation. Insufficient patient data makes it difficult to train deep learning models for medical applications, particularly for rare diseases. Efforts to share patient data are often hampered by legal, technological, and privacy issues. The current implementation of CWT has a major flaw in that it isn't designed to accommodate differences in sample sizes, mark distributions, resolution, and acquisition settings in training data through organizations. The authors \cite{Balachandar2020} present CWT modifications to reduce output losses caused by heterogeneity in training sample sizes and mark ranges through academic training splits, and test the effectiveness of their changes on virtual dispersed tasks for (DR) identification and irregular chest radiograph classification. This is the first research to show that data heterogeneity in training sample sizes and mark distributions across institutions can cause distributed learning models for medical imaging to perform poorly \cite{Balachandar2020}.
    
    \textbf{A federated network for translational cancer research}
    
    Obtaining adequate quantities of annotated human tissues remains a major barrier to translational cancer science, which is needed to move cancer treatment closer to precision medicine. Major new bridging infrastructures, including more functional biorepositories that connect human tissue to clinical phenotypes and outcomes, are required for advancements in cancer science and personalized medicine. Cancer researchers have been at the forefront of creating biomedical data and resource sharing consortia, but they have traditionally relied on centralized structures in which a single organization serves as a middleman between requesting researchers and participating institutions. The downside of centralization is that as the number of organizations grows, it becomes precarious. The TIES (Text Information Extraction System) Cancer Research Network was established by four cancer centers as a federated network that allows member organizations to share data and biospecimen.\cite{Jacobson2015} mentions pathology data that has been de-identified and analyzed with the TIES natural language processing framework can be accessed by member sites, resulting in a pool of rich phenotype data linked to clinical biospecimens. The possible effect of federated quests around the network on translational science can be seen in studies involving rare diseases, uncommon phenotypes, and complex biological behaviors. The network meets many main criteria, including local data and credentialing power, the inclusion of rich phenotype data, and applicability to a wide range of study goals \cite{Jacobson2015}.
    
     \textbf{Multi-site fMRI analysis}
    
    Data has a “non-rivalrous” value, which means it can be used by several parties at the same time to produce new data items or services, according to economics literature, Data pooling would have a synergistic impact. Sharing vast volumes of medical data is critical for precision medicine, with functional MRI (fMRI) data relating to certain neurological conditions or disorders being an interesting example. Deep learning models have shown to be useful in a variety of functions, including neuroimage processing. However, to successfully train a high-quality deep learning algorithm, a large volume of patient data must be gathered. The time and cost of acquiring and annotating massive fMRI datasets, for example, make it impossible to obtain a large number at a single location. The authors of the paper\cite{Li2020} use a privacy-preserving approach to solve the issue of multi-site fMRI classification. They suggest a federated learning approach to solve the problem, in which a decentralized iterative optimization algorithm is used and mutual local model weights are changed by a randomization mechanism. Overall, the findings show that using multi-site data without exchanging data can improve neuroimage analysis accuracy and help discover accurate disease-related biomarkers \cite{Li2020}.
    
    \textbf{Patch-Based Surface Morphometry for Alzheimer's Disease}
    
    In hospitals and academic centers, unprecedented rates of brain magnetic resonance imaging (MRI) currently exist. Simultaneously, the rapid advancement of software and hardware has made it scientifically possible to extract useful knowledge about the underpinnings of brain diseases such as Alzheimer's disease from these combined databases (AD). Researchers have faced significant challenges in obtaining or sharing these details due to patient safety issues, data limitations, and legal complications.  To address this issue, large-scale collaborative networks, such as the ENIGMA Consortium1, were established, which used secure meta-analyses to study data from hundreds of institutions around the world without sharing patients' scans or protected information. Seeking major causal factors that may predict/relate to health conditions or cognitive function, such as finding anatomically irregular regions in the brains of Alzheimer's Disease (AD) patients, is more interesting in brain imaging studies. As a result, the authors suggest a novel federated feature selection scheme based on group lasso regression using patch-based surface morphometry features from T1-weighted brain MRI images of AD, mild cognitive impairment (MCI), and stable elderly test subjects. By deliberately choosing (and visualizing) core functions, their work generalizes and enriches federated learning science. The method can discover new important features to be used as imaging biomarkers of MCI and AD by expanding access to information from large imaging datasets \cite{Wu2020}.
    
     \textbf{FADL: Federated-Autonomous Deep Learning for Distributed Electronic Health Record}
    
    Data from electronic health records (EHRs), patient-generated health data from mobile devices, and other health-related information are useful for optimizing health outcomes, especially in precision medicine. Healthcare records are kept in various locations and data silos, such as clinics, pharmacies, payors, and mobile computers. Traditionally, healthcare data are disseminated through several locations and consolidated in a network for review. However, due to stringent rules and the sensitivity of the results, healthcare data transfers are complicated. These impediments not only make data usage costly, but also slow down knowledge delivery in healthcare, where timely changes are often needed. Using ICU data from 58 independent hospitals, the authors\cite{liu2018fadlfederatedautonomous} demonstrate that machine learning models used to forecast patient mortality can be trained effectively without taking health data out of silos using a distributed machine learning approach. They suggest a new approach called Federated-Autonomous Deep Learning (FADL), which trains a portion of the algorithm using data from all data sources in a distributed manner and another portion using data from individual data sources \cite{liu2018fadlfederatedautonomous}.

     \textbf{Clinical trial protocol optimization}
    
    Clinical science is a time-consuming, labor-intensive, and expensive undertaking. Specific challenges associated with these bottlenecks include problems assessing patient demographics, determining eligible patients for enrollment, optimizing procedures, manual and inefficient data collection, data source reliability, and the difficulty of recognizing and monitoring infrequent adverse incidents. Furthermore, there are workflow problems and bottlenecks that hinder clinical trial behavior, such as suboptimal research design, slow and lengthy patient registration, site selection, and procedure optimization, all of which contribute to time and cost requirements. Some of these problems, such as protocol optimization and patient selection, can be mitigated by prudent re-use of data found in Electronic Health Records (EHRs). The growing use of EHRs in Europe and elsewhere provides a large, rich, and highly important pool of health data that has the potential to enhance clinical trial delivery. This data may be used to assess clinical trial viability using computable representations of the parameters, improve patient identification, clinical trial execution, and adverse effect monitoring. The authors in \cite{Claerhout2019} investigated the concerns by examining the Inclusion and Exclusion (I/E) criteria of 23 completed trials in a variety of therapeutic areas that were sponsored by seven pharmaceutical companies to determine the proportion of I/E criteria that could be represented in a computable format and the right to query hospital EHRs to determine the number of currently qualifying patients correctly \cite{Claerhout2019}.

    \textbf{Heart Disease predictions from Electronic Health Records}
    
    In the age of "big data," computationally efficient and privacy-aware solutions for large-scale machine learning problems are critical, especially in the healthcare context, where large volumes of data are processed in several locations and owned by various institutions. The authors here discuss three issues concerning healthcare data: (1) data exist in several places (e.g., clinics, physicians' offices, home-based computers, patients' smartphones); (2) data access is increasing, necessitating the use of scalable frameworks; and (3) aggregating data in a centralized database is infeasible or impractical due to size and/or data protection issues. Based on their medical histories as outlined in their Electronic Health Records, the authors create a distributed (federated) approach to forecast hospitalizations for patients with heart diseases within a target year (EHRs). They devise a federated optimization scheme (cPDS) to address the sparse Support Vector Machine problem. they apply their latest approach to a dataset of de-identified Electronic Cardiac Records from the Boston Medical Center, which includes patients with heart disorders \cite{Brisimi2018}.

\subsection{\textbf{Clinical Workflow}}

Clinical workflow is often conducted to optimize consistency in the workflow, analyze current frameworks, research a process and its implementation, and so on. The health care applications mentioned below conduct or include a clinical workflow on a specific disease, analysis on drug sensitivity, an EHR linking platform, and cloud-based output of federated learning on EHR's obtained from two healthcare systems to predict the risks of diseases linked to tobacco and radon.

\textbf{FedMed: A Federated Learning Framework for Language Modelling}
    
    Society is entering a smart age, with the latest breakthroughs of the modern technological revolution—Industry 4.0 and Internet of Things (IoT) technologies—where all items are enclosed with a network of interconnectivity and automation through intelligent digital technique. Meanwhile, edging systems are flooded with heterogeneous data, ranging from real-time sensor activity logs to consumer data. During the migration process, however, data is quickly attacked and poisoned. This increases the difficulty of machine learning. The emergence of federated learning techniques, as well as the obstacles and risks it poses, has occurred in recent years. Traditional FL strategies depend on averaging aggregation or don't take into account connectivity costs. The authors suggest a novel Federated Mediation (FedMed) paradigm with adaptive aggregation, mediation reward system, and topK strategy to solve concept aggregation and coordination costs in federated language modeling. Perplexity and contact rounds are used to test the results. Three datasets are used in the experiments (i.e., Penn Treebank, WikiText-2, and Yelp) \cite{Wu2020}
    
    \textbf{Federated learning on clinical benchmark data}
    
    FL may be used to address privacy concerns and reduce the possibility of a data violation in clinical records so data transfer and centralization are not necessary. Since medical data is among the most vulnerable forms of personal data, privacy protection is especially important for medical data processing. De-identification techniques have traditionally been used to protect patients' privacy. A performance evaluation using federated learning on clinical benchmark data was performed. The Modified National Institute of Standards and Technology (MNIST) dataset, Medical Information Mart for Intensive Care-III (MIMIC-III) dataset, and PhysioNet Electrocardiogram (ECG) dataset were used in a federated learning analysis. By changing the MNIST, MIMIC-III, and ECG datasets, they also validated FL in environments that simulate real-world data distributions \cite{Lee2020}.
    
    \textbf{Identifying potential risk variants in ankylosing spondylitis}
    
    Genome-wide association studies (GWAS) have been common for detecting possible risk variants in a variety of diseases. Large sample size is usually needed for a statistically significant GWAS to identify disease-associated single nucleotide polymorphisms (SNPs). A single institution, on the other hand, normally only has a small number of samples. As a result, cross-institutional collaboration is expected to maximize sample size and statistical capacity. However, cross-institutional collaborations present serious problems, one of which is data protection. While data sharing within a broad network can greatly support biomedical science, it also poses potential threats to data privacy due to the exchanging of personal information about individuals. The consequences of patient private information leaks include, but are not limited to, workplace discrimination, denial of benefits, higher insurance premiums, and so on. To address the problems of data privacy due to the exchange of personal information, the authors suggest a novel privacy-preserving federated GWAS architecture (iPRIVATES). iPRIVATES, which includes privacy-preserving federated processing, allows various organizations to collaborate on GWAS analysis without disclosing patient-level genotyping results \cite{Wu2021}.
    
    \textbf{Drug sensitivity prediction}
    
    Users with a personalized recommendation framework face a conundrum: learning from data will boost recommendations, but only when other users can share their private anonymization. Good personalized forecasts are critical in precision medicine, but the genetic knowledge from which the predictions are based is often especially vulnerable since it clearly describes the patients and therefore cannot be conveniently anonymized. Genomics is a critical domain for privacy-aware modeling, especially in precision medicine. Many people like to keep their and their descendants' genomes secret, but basic anonymization is insufficient since a genome is inherently recognizable. As a result, the hospital or clinic that holds the genomic data must be very vigilant about privacy concerns when disclosing the genomic data, even though the data is required and helpful for potential diagnosis and care decisions. Even with moderate-sized data, the proposed differentially private regression approach \cite{Honkela2017} combines the theoretical appeal and asymptotic efficiency with good prediction accuracy. Under comparatively strict differential privacy assurances, their approach exceeds the predictive precision of state-of-the-art non-private lasso regression with just 4x more tests \cite{Honkela2017}.
    
     \textbf{Integration of Medical Data}
    
    Medical evidence is often maintained by many organizations. However, comprehensive evaluations can necessitate the integration of these datasets without jeopardizing patient or commercial privacy. An effective privacy-preserving algorithm, Multiparty Private Set Intersection (MPSI), computes the intersection of several private datasets. Multiparty Private Set Intersection (MPSI), computes the intersection of several private datasets using this, the authors suggest a functional MPSI with the following characteristics: the size of each party's datasets is independent of that of the other parties, and the statistical complexity for each party is independent of the number of parties \cite{Miyaji2016}.
    
    \textbf{Design and implementation of EHRs}
    
    Electronic clinical evidence has increased dramatically as a result of healthcare laws and government incentives encouraging the use of Electronic Health Records (EHRs). As a result, academics and public health authorities have shown a growing interest in data linkage that can be used in cross-site health studies. Linking EHR data through healthcare agencies, on the other hand, necessitates striking a balance between data accessibility and privacy. Some regions have adopted health information exchange (HIE) programs to provide healthcare facilities with up-to-date clinical data on patients through hospitals for better and more organized treatment, as part of inter-institutional arrangements for the sharing of PHI. However, owing to concerns about reliability, anonymity, and protection, as well as other problems, organizational HIEs exchanging patient-level identifiers, are still not commonly used. The authors define a real-world implementation of a software framework (Distributed Common Identity for the Integration of Regional Health Data – DCIFIRHD) that uses a structured and distributed encryption algorithm to conduct stable, cross-site aggregation and linking of EHR data for analysis. As part of the HealthLNK study initiative, they applied the application in a major metropolitan area (Chicago, IL, USA), aggregating over 5 million patients' clinical records through six healthcare institutions \cite{Kho2015}.

\textbf{Cloud-based FL}

From diagnosis to medication decisions, machine learning (ML) models can improve health care. However, ML model generalizability is jeopardized due to a lack of adequate heterogeneous data due to patient privacy concerns. When compared to a simulation in a single organization, heterogeneous data from multiple centers increased model accuracy. Furthermore, cloud systems come equipped with the required tools and security measures to support federated learning deployments. The amount and quality of data used to train an ML algorithm are extremely important, particularly for more complex models. The availability of diverse multidimensional patient data sets in the age of precision medicine necessitates greater population surveys for generalization. Furthermore, data shortage of underrepresented communities may contribute to prejudices if training data does not adequately portray these populations' characteristics. Quality of healthcare data and algorithmic problems are also well-known ML roadblocks. To assess the federated machine learning solution, the authors use electronic health record data from two academic medical centers on a Microsoft Azure Cloud Databricks network to test various federated learning applications in both a virtual and real-world setting. Using data from two healthcare systems' electronic health records (EHRs), they trained machine learning models to forecast the risks of diseases linked to tobacco and radon \cite{Rajendran2021}.

\textbf{Clinical decision support system}

Clinical decision support systems (CDSS) are computer-based applications designed to improve patient care and healthcare delivery by assisting clinicians in analyzing health information better and enhancing medical decisions. \cite{Sutton2020AnOO,Kawamoto2005ImprovingCP} A CDSS uses knowledge management to get clinical insights from the patient and health-related data based on multiple factors. Clinical decision support systems assist the clinicians at the time of care and help to make a better care plan \cite{bright2012effect,al2013clinical}.

CDSS comprises three main components, including Base, inference engine, and Communication Mechanism. The base can be classified into two types of systems called knowledge-based and non-knowledge-based.

Knowledge-based support systems are defined by well-established rules(IF-THEN statements) that determine 'what is true?'.
Support systems without a knowledge base, on the other hand, answer the question "what to do?" advising on the next steps for treatments, which drug to prescribe, etc. These systems still require a data source and use artificial intelligence (AI), statistical pattern recognition, or machine learning (ML) to make medical decisions and provide recommendations \cite{gultepe2014vital,valdes2017clinical,baig2016machine}.

A large amount of data is often required to train an AI system, but clinical data's varied and sensitive aspects present a barrier to the standard centralized network. The author of the paper \cite{Thwal2021AttentionOP} proposed a deep learning-based clinical decision support system trained and managed under a federated learning paradigm to solve the data and privacy challenges. The paper utilized a novel strategy to ensure the safety of patient privacy and overcome the risk of cyberattacks while enabling large-scale clinical data mining.

\begin{longtblr}[
caption = {Summary of existing studies on Deployment of ML and FL in Healthcare since 2016},
  label = {Tab:dcnnarchitectures1},
                ]
                {
    rowhead=1,
    hline{1,2,Z} = 1pt, hline{3-Y},
    colsep = 3pt,
    colspec = {@{} X[j,h] Q[c,m] Q[c,m] Q[c,m] Q[c,m] Q[c,m] Q[c,m] @{}},
    rows = {font=\footnotesize},
    row{1} = {font=\footnotesize\bfseries},
                }
Title   & Ref   & ML Model & Data Type & Year  & \#Hospitals/clients  & \#Samples      \\
%

Distributed learning: Developing a predictive model based on data from multiple hospitals without data leaving the hospital – A real life proof of concept
& \cite{Jochems2016}
& Bayesian network
& EHR
& 2016
& 5
& 287 \\

Developing and Validating a Survival Prediction Model for NSCLC Patients Through Distributed Learning Across 3 Countries
& \cite{Jochems2017DevelopingAV} 
& SVM
& EHR
& 2017
& 3
& 894 \\

Patient clustering improves efficiency of federated machine learning to predict mortality and hospital stay time using distributed electronic medical records
& \cite{Huang2019} 
& k-means
& EHR
& 2019
& 208
& 200,859 \\

Distributed learning on 20 000+ lung cancer patients – The Personal Health Train
& \cite{Deist2020} 
& Logistic regression
& EHR
& 2020
& 8
& 23,203 \\

Federated Learning of Electronic Health Records Improves Mortality Prediction in Patients Hospitalized with COVID-19
& \cite{Vaid2021} 
& Federated MLP,LASSO 
& EHR & 2020 
& 5 
& 4029 \\

Joint Imaging Platform for Federated Clinical Data Analytics
& \cite{Scherer2020} 
& CNN based organ segmentation
& Images
& 2020
& 10
& - \\

Stochastic Channel-Based Federated Learning With Neural Network Pruning for Medical Data Privacy Preservation: Model Development and Experimental Validation
& \cite{Shao2020} 
& MLP
& EHR
& 2020
& 5
& 30,760 \\

Federated learning in medicine: facilitating multi-institutional collaborations without sharing patient data
& \cite{Sheller2020} 
& U-Net
& Images
& 2020
& 13
& 352 \\

Federated semi-supervised learning for COVID region segmentation in chest CT using multi-national data from China, Italy, Japan
& \cite{Yang2021} 
& Fully Convolutional Network (FCN) 
& Images 
& 2021
& 3
& 1704 \\

Real-Time Electronic Health Record Mortality Prediction During the COVID-19 Pandemic: A Prospective Cohort Study
& \cite{Sottile2021} 
& Stacked regression model
& EHR
& 2021
& 12
& 28,538 \\

Federated Learning used for predicting outcomes in SARS-COV-2 patients
& \cite{Flores2021} 
& ResNet-34
& Images
& 2021
& 20
& 16,148 \\

Cloud-Based Federated Learning Implementation Across Medical Centers
& \cite{Rajendran2021} 
& ANN / Logistic regression
& EHR
& 2021
& 2
& 10,000 \\

Federated learning improves site performance in multicenter deep learning without data sharing
& \cite{Sarma2021} 
& 3D Anisotropic Hybrid Network
& Images
& 2021
& 3
& 300 \\

\end{longtblr}

\section{\textbf{Tools}}

\begin{figure}[h]
  \centering
  \includegraphics[width=8cm]{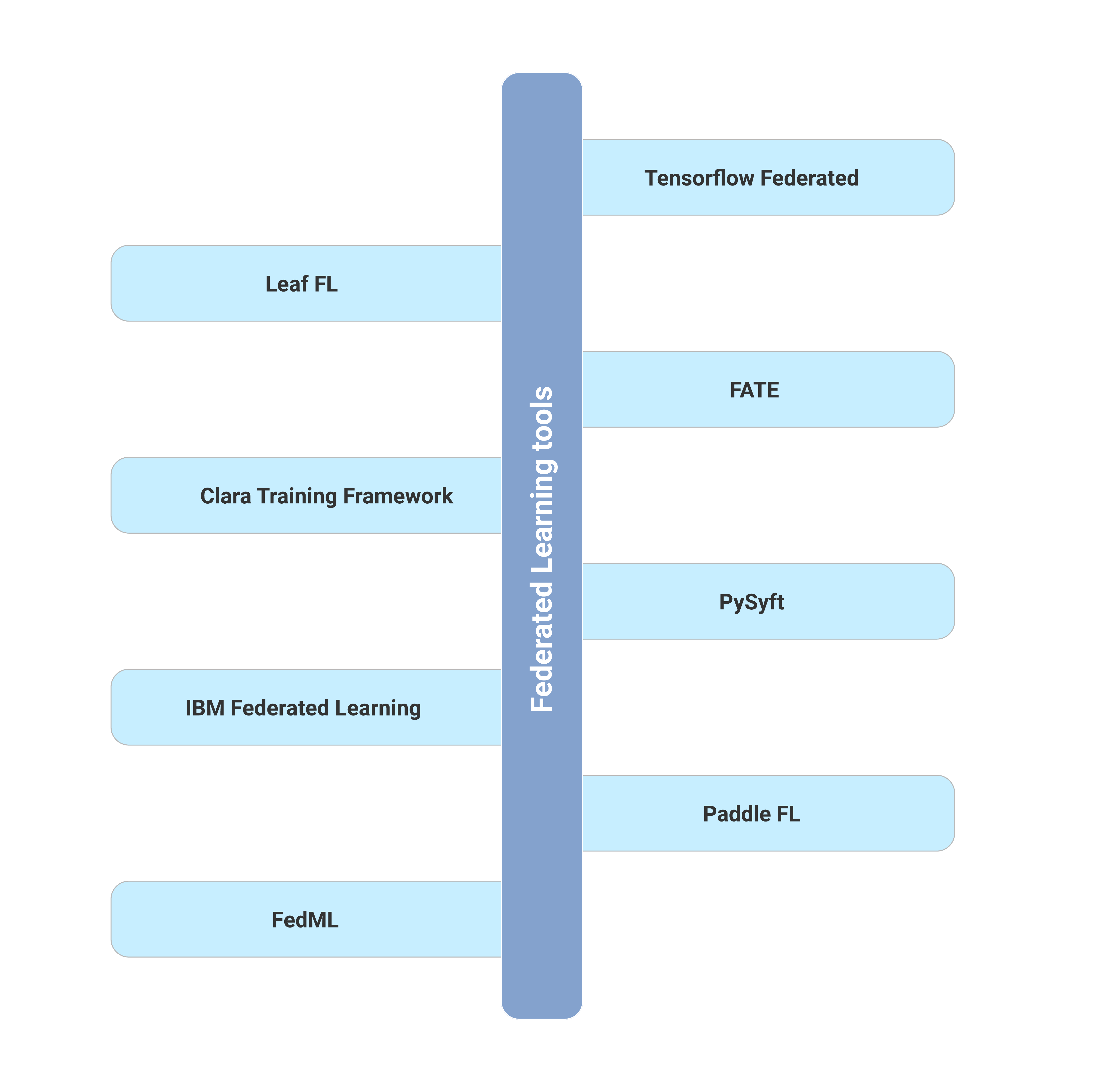}
  \caption{Tools of federated learning}
  \Description{A woman and a girl in white dresses sit in an open car.}
\end{figure}

\subsection{\textbf{TensorFlow Federated}}
TFF \cite{TFL} is an open-source framework to perform machine learning and computations on localized information. TFF allows developers to simulate federated learning algorithms on the models and data. TFF’s interfaces are unionized in 2 layers. Federate Core(FC) and Federate Learning (FL) APIs.Federated Core API is the foundation layer upon which is built the Federated Learning API. This is a strong functional programming environment that involves a combination of novel federated learning algorithms and TensorFlow with distributing higher-revelations operators at the core of the system. Whereas Federated learning API is at a higher-level interface. This API allows developers to use the existing FL algorithms (training and evaluation) in their models.
    Developers can incorporate their functions and interfaces within the federated heart. FC, as a Python package, includes Python interfaces that can be used to create new Python features. It supports several styles, including Tensor types, sequence types, tuple types, and function types, to make it simple to use, particularly for developers familiar with TensorFlow. It supports a wide range of federated operators, including federated sum, federated minimize, and federated broadcast.
    
    TFF currently only supports FedAvg and does not provide any privacy mechanisms. It can now only be deployed on a single computer, with the federated setting applied by simulation.

\subsection{\textbf{Federated AI Technology Enabler (FATE) }}

    Webank’s AI Department initiated an open-source project called FATE \cite{Fate}. This framework provides a secure computing framework supporting the AI Ecosystem. It uses privacy-preserving techniques like Homomorphic encryption and Secured Multi-Party Computation. It also supports ML algorithms like Logistic regression, Deep learning, tree-based algorithms, and transfer learning. EggRoll, FederatedML, FATE-Flow, FATE-Serving, FATE-Board, and KubeFATE are its six main modules.

    The federated algorithms and stable protocols are used in FederatedML. It currently supports training a variety of machine learning models, including NNs, GBDTs, and logistic regression, in both horizontal and vertical federated settings. To ensure anonymity, it also incorporates safe multi-party computing and homomorphic encryption. To run an FL algorithm, users simply set the parameters. FATE also includes comprehensive documentation of how to deploy and use it.

Practitioners must change FATE's source code to incorporate their federated algorithms because FATE has algorithm-level interfaces. Non-expert consumers would find this difficult.

\subsection{\textbf{PySyft}}

    PySyft \cite{ryffel2018generic}  is a secure and private deep learning library. It is a library for answering questions using data you cannot see. It uses privacy-preserving techniques like Differential Privacy, Homomorphic Encryption, and multi-party computation with deep learning frameworks like Pytorch, and Tensorflow.

    Both PyTorch and TensorFlow can be used with PySyft. It can be installed on a single computer or several computers, with the WebSocket API used to communicate between clients.

    Though PySyft offers several tutorials, there is no comprehensive documentation on the system's interfaces or architecture. PySyft does not support a diversified computing paradigm like on-device training on Mobile or IoT.

\subsection{\textbf{Leaf}}

    LEAF \cite{Leaf} is a benchmarking framework for studying in federated settings, with programs consisting of federated learning, multi-task mastering, meta-learning, and on-tool learning.

    It includes six databases that cover a variety of topics, such as image recognition, emotion analysis, and next-character prediction. A collection of utilities is given to split datasets into separate parties in an IID or non-IID manner. A reference implementation is also presented for each dataset to explain how to use the dataset in a training phase.

    Leaf databases have enough clients to model cross-device FL scenarios, but they may be too limited for questions where size is especially relevant. It only supports standardized algorithm implementations such as Fed Avg and does not support decentralized federated learning, split learning, and vertical federated learning.

\subsection{\textbf{Paddle FL}}

    Paddle FL \cite{PaddleFL} provides applications in Natural Language Processing, Computer Vision, and recommendation systems. Paddle FL allowing the deployment of federated learning systems in the form of distributed clusters at a large scale has been proven to be of great benefit to developers. PaddleFL is an open-source framework based on PaddlePaddle. 

    FL methods, user-specified models and algorithms, distributed training setup, and FL task generator are all included in the compile time. The horizontal FL algorithms, such as FedAvg, are among the FL techniques. Users are allowed to create their models and training algorithms in addition to the FL techniques offered.
    
    PaddleFL is still in its early stages of production, and the documentation and samples are lacking. It also lacks in performance with standardized benchmarks such as Model DNN (eg: ResNet) and vertical federated learning. Paddle FL does not involve a flexible and generic API design with topology customization and flexible message flow.

\subsection{\textbf{Clara Training Framework}}

Clara Train SDK \cite{Clara} is a domain-optimized developer software framework that consists of APIs for AI-Assisted Annotation. This allows any clinical viewer to be AI successful and allows a TensorFlow-based framework with pre-skilled models to begin AI improvement with strategies including Transfer Learning, Federated Learning, and AutoML.

Developers may use the Clara Train SDK's configurable MMAR (Medical Model ARchive) function to carry their models and components to conduct Federated Learning, as well as monitor whether the local training is run on a single GPU or multiple GPUs.

\subsection{\textbf{Fed ML}}
FedML \cite{chaoyanghe2020fedml} serves as a tool for federated learning as well as a forum for FL benchmarking. Its core structure is split into two layers as an FL system. On-device training for IoT and mobile computers, distributed computing, and single-machine emulation are all supported. FedML also embraces a variety of algorithms, prototypes, and databases for study variability (e.g., decentralized learning, vertical FL, and split learning).

FedML offers a simulation environment for a wide range of hardware specifications while supporting three computing paradigms \cite{chaoyanghe2020fedml}: standalone simulation, distributed computing, and on-device training.

\subsection{\textbf{IBM Federated Learning}}
A python framework library for distributed machine learning processes in an enterprise environment \cite{IBMFL}.IBM Federated Learning focuses on enterprise environments where safe rollout, failure tolerance, and fast model specification are critical; these must make use of existing machine learning libraries that enable enterprise users to access a robust collection of state-of-the-art algorithms without learning new languages. Enterprise professionals will be able to easily implement federated learning using IBM Federated Learning.

Apart from neural networks and decision trees, IBM Federated Learning facilitates the learning of a variety of machine learning models such as multi-class classification,regression,linear classifiers and adaptation of XGBoost. IBM Federated Learning also gives you the ability to apply differential privacy to a variety of models, from basic Naive Bayes to more sophisticated differential privacy structures like those used in neural networks.

\section{\textbf{CONCLUSION \& FUTURE DIRECTIONS}}
Federated learning is a learning paradigm where machine learning models are trained at the edge. It was originally designed for a variety of domains, including mobile and edge device use cases, but it has recently acquired popularity in healthcare applications.

The development of federated learning systems for healthcare has sparked a lot of interest from both industry and academics. As a result, a comprehensive overview and summary of existing FLSs in the healthcare domain is required.

Not all technological concerns have been solved, and FL will undoubtedly be a focus of study in the coming decade.
Even though 5G networks are not yet widely available and commercially implemented worldwide, significant research and development efforts have been directed toward future 6G wireless systems.\cite{Alwis2021SurveyO6}

Future research will focus on incorporating FL functionalities into future 5G/6G medical devices, how to use 6G devices, such as intelligent implants and wearables, for large-scale FL-based healthcare, and what new healthcare services 6G enables.
Future e-health services, for example, will be enhanced by AI and FL capabilities, improving patient quality of life and lowering hospitalizations \cite{Mucchi2020How6T}.

In this article, we presented an overview of federated learning in the healthcare industry. We conducted a comprehensive survey of recent work in the healthcare sector, with a focus on federated settings. We also discussed how to employ federated learning in healthcare, as well as the methods, applications, and issues that come with it.
\bibliographystyle{ACM-Reference-Format}
\bibliography{sample-base}

\end{document}